\pdfoutput=1

\documentclass[11pt]{article}

\usepackage[final]{acl}

\usepackage{times}
\usepackage{latexsym}

\usepackage[T1]{fontenc}

\usepackage[utf8]{inputenc}

\usepackage{microtype}

\usepackage{inconsolata}

\usepackage{graphicx}

\usepackage[utf8]{inputenc}
\usepackage{textcomp}
\DeclareUnicodeCharacter{266B}{\textmusicalnote}
\usepackage{microtype}

\usepackage{inconsolata}

\usepackage{multicol}
\usepackage{booktabs}
\usepackage{makecell}
\usepackage{amsmath, amssymb}
\usepackage{multirow}
\usepackage{mathtools}
\usepackage{xcolor}
\usepackage{enumitem}
\usepackage{float}
\usepackage{graphicx}
\usepackage{upgreek}
\usepackage{seqsplit}
\usepackage{color,soul}
\usepackage{arydshln}
\usepackage{placeins}
\usepackage{pifont}
\usepackage{amssymb}
\usepackage{url}
\usepackage{bbm}
\usepackage{array}
\usepackage[table,x11names]{xcolor}
\usepackage{colortbl}
\usepackage{xspace}
\usepackage{comment}

\usepackage{amsmath,amsfonts,bm}

\def\eqref#1{equation~\ref{#1}}

\def\1{\bm{1}}

\def\vd{{\bm{d}}}

\def\vq{{\bm{q}}}
\def\vr{{\bm{r}}}

\def\mR{{\bm{R}}}

\DeclareMathAlphabet{\mathsfit}{\encodingdefault}{\sfdefault}{m}{sl}
\SetMathAlphabet{\mathsfit}{bold}{\encodingdefault}{\sfdefault}{bx}{n}

\usepackage{caption}
\usepackage{subcaption}
\title{Database-Augmented Query Representation for Information Retrieval}

\author{Soyeong Jeong$^1$
        \; Jinheon Baek$^1$
        \; Sukmin Cho$^1$
        \; Sung Ju Hwang$^{1, 2}$
        \; Jong C. Park$^1$\thanks{\hspace{0.2cm}Corresponding author} \\
        KAIST$^{1}$ \;\; DeepAuto.ai$^{2}$ \\
       \texttt{\{starsuzi, jinheon.baek, nelllpic, sungju.hwang, jongpark\}@kaist.ac.kr}}

\begin{document}
\maketitle

\begin{abstract}
Information retrieval models that aim to search for documents relevant to a query have shown multiple successes, which have been applied to diverse tasks. Yet, the query from the user is oftentimes short, which challenges the retrievers to correctly fetch relevant documents. To tackle this, previous studies have proposed expanding the query with a couple of additional (user-related) features related to it. However, they may be suboptimal to effectively augment the query, and there is plenty of other information available to augment it in a relational database. Motivated by this fact, we present a novel retrieval framework called Database-Augmented Query representation (DAQu), which augments the original query with various (query-related) metadata across multiple tables. In addition, as the number of features in the metadata can be very large and there is no order among them, we encode them with the graph-based set-encoding strategy, which considers hierarchies of features in the database without order. We validate our DAQu in diverse retrieval scenarios, demonstrating that it significantly enhances overall retrieval performance over relevant baselines. Our code is available at \url{https://github.com/starsuzi/DAQu}.
\end{abstract}

\section{Introduction}

Information Retrieval (IR) is the task of fetching query-relevant documents from a large corpus. Traditional approaches have focused on sparse retrieval, which searches for documents that yield the highest lexical match with the query~\cite{bm25}. Recently, neural language models have led to the introduction of dense retrieval models, which represent both the query and the document in a learnable latent space and then calculate their similarity on it~\cite{dpr, contriever, bge-m3}.
Notably, these IR methods have gained much attention in the era of Large Language Models (LLMs), due to their ability to assist LLMs help generating accurate answers with evolving knowledge from an external source~\cite{das, adaptiverag}.

\begin{figure*}
    \centering
    \includegraphics[width=0.975\linewidth]{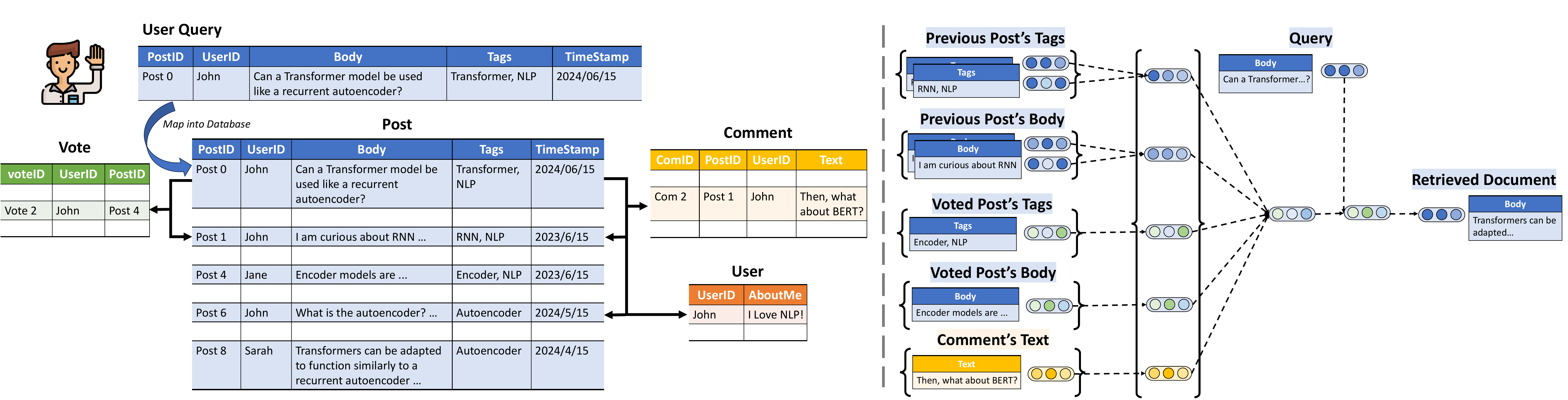}
    \vspace{-0.025in}
    \caption{A conceptual illustration of our proposed DAQu, which shows a link among multiple tables for the given query (Left) and visualizes a graph-based set-encoding strategy that encodes metadata hierarchically for query augmentation (Right).}
    \label{fig:concept}
    \vspace{-0.075in}
\end{figure*}

Despite such a huge advantage of IR in NLP, it faces a critical challenge that information captured in a query itself is oftentimes not sufficient to retrieve its relevant documents, due to the scarcity of information within its (shorter) text. To tackle this challenge, previous work has focused on enriching representations of queries or documents by expanding them with additional texts or augmenting their representation spaces~\cite{dar,jagerman2023query,dragon}. Yet, despite their improvements, those approaches are still limited in that they rely on the capability of models themselves (e.g., LLMs) used during augmentation, though there can be external knowledge sources (for augmentation) associated with the user query (such as the user's purchase history for shopping). 

While some other work has considered these additional sources, enhancing the representation of queries with them, they leverage only a single source of information stores, especially the one specific to the user (who issues the query)~\cite{amazonqa,DBLP:conf/sigir/ZhangDL20,Deng2021TowardPA,DBLP:conf/vldb/BussMTT0L23}. However, in the real-world, data (including queries) is usually mapped into the database and linked to other data within it, which means that plenty of information that can be potentially used for query enrichment is available on the relational database~\cite{relbench}. 
For example, online platforms like e-commerce often use relational databases to store and link structured information such as user profiles, purchase histories, and prior interactions. Similarly, healthcare databases connect patient queries to records like medical histories and lab results, while travel databases associate queries with itineraries and customer profiles.

Therefore, in this work, we introduce a novel IR paradigm, Database-Augmented Query representation (DAQu), which augments representations of queries by searching for and connecting their associated information across multiple tables within the relational database. As shown in Figure~\ref{fig:concept}, consider the task of identifying the answer post that the user would most likely to vote as the best. In this scenario, we can not only represent the query with its own information but also with its relevant information within and across the multiple tables. Specifically, we can use metadata in the same table, such as its tags, but also metadata spread over other multiple tables, which include user-specific information, such as previous posts, answers (that they voted for), bios, and badges (that they earned). For example, given the question from the user, “Can a Transformer model be used like a recurrent autoencoder?”, user tags like “Transformer” and “Autoencoder” can emphasize the focus on these specific concepts. Further, the user’s past questions about “RNNs” and “Autoencoders” reveal an existing familiarity with these topics, while the Vote table highlights which answers the user has previously favored, offering further insight into their preferences. 
However, the volume of these metadata can be extremely large, and simply expanding the query with additional terms in the metadata (as done in existing query expansion work~\cite{amazonqa, Deng2021TowardPA}) is not feasible due to the limited context length of LMs. Moreover, since there is no inherent order for the elements in the metadata, the query augmentation approach should ensure order invariance when incorporating this information.

To this end, we further propose to encode various query-related metadata within and across multiple tables over the relational database, based on a graph set-encoding scheme. Specifically, this strategy models metadata for query expansion as a two-layer hierarchical graph structure, and, within this structure, the first layer aggregates query-related elements (cells) within each column into a column-level representation, and next the second layer aggregates these column-level representations into a query-level representation. For example, consider a query from the Stack Exchange dataset in Figure~\ref{fig:concept}, which is linked to metadata such as the user's profile, previous posts, and associated tags. Then, each individual attribute (e.g., a tag, a user bio, and a body of the previous post) is first encoded independently. After that, within each column (e.g., tags), these encoded attributes are aggregated to create the column-level representation (e.g., all tags combined into a single vector). Lastly, all column-level representations (for tags, user bios, and previous post content) are aggregated into the final query-level metadata representation that is used to enrich the original query representation. It is worth noting that those two-layer structures (aggregation on column- and query-level) can be viewed as a two-layer graph neural network~\cite{GCN, MPNN} since the first layer models interactions within columns (i.e., intra-column relationships) and the second layer models interactions across columns (inter-column relationships).

We then validate our DAQu on seven different retrieval tasks designed with multiple databases of Stack Exchange, Amazon Product Catalog, and H\&M~\cite{relbench, relbench/new}. The experimental results show significant improvements of our DAQu in retrieval performance compared to other query augmentation baselines across diverse scenarios. Moreover, we demonstrate that the graph set-encoding technique operationalized in our DAQu effectively represents metadata, enhancing the representations of queries for retrieval.

\section{Related Work}

\paragraph{Retrieval}
In response to a query from a user, the retrieval task is to search for the most relevant documents from a large corpus (such as Wikipedia)~\cite{retriever_survey}. Typically, it can be performed with two types of models: sparse and dense retrievers. Specifically, sparse retrievers such as TF-IDF or BM25~\cite{bm25} represent the query and document based on their terms and frequencies in a sparse vector space, whereas dense retrievers use a trainable dense vector space to embed the query and document usually with language models~\cite{dpr, contriever, bge-m3}. Recently, due to the limitation of sparse retrievers that are vulnerable to the vocabulary mismatch problem, dense retrieval is widely selected as a default choice and many advancements have been made on it~\cite{rag_llm_survey}. For example, DPR~\cite{dpr} is a supervised dense retriever with a dual-encoder architecture that is trained discriminatively on the labeled pair of a query and its relevant documents to achieve higher similarity scores than the pair of the query-irrelevant documents. Also, Contriever~\cite{contriever} utilizes a self-supervised learning strategy, which generates its training samples by creating positive pairs from query-related contexts within and across documents, rather than relying on explicitly annotated data. Yet, using only the information within a query for retrieval can be suboptimal, due to the scarcity of information on it.

\paragraph{Query Augmentation for Retrieval}
Some studies have proposed augmenting the original query with additional information to enhance the retrieval performance~\cite{DBLP:journals/csur/CarpinetoR12, DBLP:journals/ipm/AzadD19}. Specifically, traditional augmentation methods have focused on utilizing a lexical knowledge base such as the WordNet~\cite{wordnet} to expand the original queries~\cite{DBLP:journals/ipm/BhogalMS07, 5231693}. In addition, some other work has implemented statistical models such as RM3~\cite{rm3}, which add new terms to the query extracted from the top documents in the initial search results and then adjust their weights based on their importance~\cite{DBLP:conf/sigir/LavrenkoC01, DBLP:conf/trec/JaleelACDLLSW04, DBLP:conf/cikm/LvZ09a}. However, they have been shown to be not very effective and, in some cases, even degraded the performance~\cite{Nogueira2019document, jeong-etal-2021-unsupervised}. Therefore, recent work has turned to leveraging neural models to extract or generate query-relevant terms and then append such terms to the original query~\cite{DBLP:journals/isci/EspositoDMPF20, DBLP:conf/emnlp/ZhengHHH0Y20, DBLP:conf/acl/MaoHLSG0C20}, or exploiting multiple fields within the document itself~\cite{multi-field-adaptive-retrieval}. Moreover, some studies further use recent LLMs to utilize their remarkable capabilities in generating such terms~\cite{DBLP:conf/emnlp/WangYW23, DBLP:conf/emnlp/ShaoGSHDC23,DBLP:conf/vldb/BussMTT0L23, jagerman2023query, 10448015, DBLP:conf/ecir/DholeA24, DBLP:journals/corr/abs-2410-13765}. However, despite the fact that the query is represented and leveraged in the latent space with the recent dense retrievers, existing work focuses on explicitly expanding its text (instead of manipulating this query representation for augmentation). This approach may be problematic if there is a significant amount of data available to augment the query across multiple relational tables over the database. 

\paragraph{Retrieval with Database}
A natural way to store a collection of data is to use a relational database, that is designed to effectively manage, retrieve, and manipulate data for various applications~\cite{MIMIC-III,relbench}. To utilize the data in the database, the task of retrieving the tabular structures and the information in them has received much attention. Specifically, some studies have developed the approach to retrieve tables themselves (relevant to the given query) from a large table corpus~\cite{DBLP:conf/naacl/HerzigMKE21, wang-etal-2022-table}. Some other work extends this approach, extracting or generating the answer for the query from the retrieved tables~\cite{ctrl, trag, DBLP:conf/acl/LinBBGI23a}. However, since some real-world questions require multiple tables, recent studies have made further progress, proposing to incorporate multiple tables during retrieval~\cite{Open-WikiTable, multi_table_retrieval} or reading the tables~\cite{multitableqa}. However, unlike all the aforementioned work that has focused on retrieving the tables themselves and finding relevant cells within them, our work is completely different, which aims to effectively handle the query for document retrieval by using the query-related information spread across multiple tables, to augment the representation of the query.

\section{Method}

\subsection{Preliminaries}
We begin with preliminaries, providing formal descriptions of the retrieval and query augmentation.

\paragraph{Dense Retrieval}
Let us define the query as $q$ and its relevant document as $d \in \mathcal{D}$, where $\mathcal{D}$ is a corpus. To operationalize retrieval, we should be able to calculate the similarity between $q$ and $d$: $f(q, d)$, where $f$ is a scoring function. Following the bi-encoder architecture for dense retrieval, we obtain the similarity by representing the query and document with encoders $\texttt{Enc}_q$ and $\texttt{Enc}_d$ parameterized by $\theta_q$ and $\theta_d$, respectively, formalized as follows: 
\begin{equation}
\begin{aligned}
    &f(q, d) = \text{sim}(\vq, \vd), \\
    &\vq = \texttt{Enc}_q(q; \theta_q) \quad \text{and} \quad \vd = \texttt{Enc}_d(d; \theta_d),
\end{aligned}
\label{eq:similarity}
\end{equation}
where $\vq$ and $\vd$ are the query and document representations, respectively. In addition, $\text{sim}$ is a similarity metric (e.g., cosine similarity). It is worth noting that the objective of the dense retrieval function $f$ is to rank the pair of query $q$ and its relevant document $d^+$ highest among all the other pairs with irrelevant documents $\{ d_i^- \}_{i=1}^N$. To reflect this, we formalize the training objective, as follows:
\begin{equation}
    l = -\log \frac{e^{f(q,d^+)}}{e^{f(q,d^+)} + \textstyle\sum_{i=1}^N e^{f(q,{d_i}^-)}}.
\label{eq:contrastive}
\end{equation}

\paragraph{Query Augmentation for Retrieval}
To improve the effectiveness of the dense retrieval (while tackling the limited contextual information within the query $q$), the textual query itself or its representation $\vq$ can be enriched by augmenting it with the information that is not present in the original $q$. In this work, to effectively incorporate diverse pieces of information into the query without their order variance, we turn to augmenting the query representation $\vq$ over the latent space, as follows:
$\tilde{\vq} = \lambda \vq  + (1-\lambda) \vq'$, 
where $\tilde{\vq}$ is the reformulated representation, $\vq'$ is the representation of the additional information helpful to enrich the original query $\vq$, and $\lambda \in [0, 1]$ is for giving weight to it.

\subsection{Database-Augmented Query~Representation}
\label{subsec:daqu}
We now introduce our Database-Augmented Query representation (DAQu) framework for IR.

\paragraph{Relational Database}
As a vast amount of information is typically stored in a relational database, we aim to augment the representations of queries with the relevant information within this database. The relational database can be defined as a set of tables: $\mathcal{T} = \{T_i\}_{i=1}^{N}$, and each table is comprised of a collection of rows $T =\{r_j\}_{j=1}^{K}$, where $N$ is the number of tables and $K$ is the number of rows. 

Note that one of the valuable characteristics of the relational database is that some rows in tables are connected with others in other tables, which facilitates relational linkages and ease of data retrieval. Formally, each row $r_i$ in the table consists of a primary key column that uniquely identifies each row within the table, (potentially) some foreign key columns that link to primary keys in other tables, and other non-key attribute columns providing additional information about the row. In other words, the relationships between primary and foreign keys connect rows across different tables, and other attribute columns store descriptive information. Formally, if a foreign key column $f$ in table $T_i$ references a primary key column $p$ in $T_j$, we can represent their relationship as $(f_i, p_j)$. Also, all such relationships between tables can be denoted as $\mathcal{L} = \{(f_i, p_j)\}_{(i, j)}$ where $\mathcal{L} \subseteq T \times T$.

For example, analogous to the Amazon database, let's assume that the table $T_{review}$ includes the primary key column \textsc{ReviewId}, the foreign key column \textsc{ProductId}, and the attribute column \textsc{Text}. Also, the table $T_{product}$ has the primary key column \textsc{ProductId} and the attribute column \textsc{Description}. Lastly, the foreign key column \textsc{ProductId} in $T_{review}$ points to the primary key column in $T_{product}$. Then, the relationships between those two tables can be represented with a pair of primary and foreign keys: $(\textsc{ProductId}_{review}, \textsc{ProductId}_{product})$.

\paragraph{Query Augmentation with Relational Database}
Recall the equation to augment the representation of the given query ($\tilde{\vq} = \lambda \vq  + (1-\lambda) \vq'$). Here, $\vq'$ is the representation that we obtain from the query-related information within the relational database, and we now turn to explain how to get this $\vq'$.

Formally, each query that the user requests can be considered as one row $r_j$ in a certain table $T_i$. For example, in the Stack Exchange dataset, the query that the user posts is stored in the table as one row: $r \in T_{post}$, where this row (query) $r$ consists of the primary key ($\textsc{PostId}$), the foreign key ($\textsc{UserId}$), and the multiple attributes (such as $\textsc{Body}$, $\textsc{Tags}$, and $\textsc{TimeStamp}$). Then, based on the following relational structure of this database: 
\begin{equation}
\begin{aligned}
    \mathcal{L} = \{ 
    & (\textsc{UserId}_{user}, \textsc{UserId}_{post}), \\
    & (\textsc{UserId}_{vote}, \textsc{UserId}_{post}), \\
    & (\textsc{PostId}_{post}, \textsc{PostId}_{comment}), ...
    \},
\end{aligned}
\label{eq:relation}
\end{equation}
the row for the query in the post table can be linked to other rows in different tables, for example, the user table, vote table, and comment table connected with \textsc{UserId} and \textsc{PostId} columns (Figure~\ref{fig:concept}). 

This relational structure of the database allows us to utilize diverse pieces of information when enriching the query representation $\vq$. Specifically, we can not only use the attributes within the columns of the row for the query (such as $\textsc{Body}$ and $\textsc{Tags}$ of the post table $T_{post}$) but also the attributes of associated rows (to the query) from different tables (such as $\textsc{AboutMe}$ of the user table $T_{user}$ associated with the column $\textsc{UserId}$). Formally, all the attributes of rows associated with and used to augment the query ($q$) can be represented as follows: 
\begin{equation}
\fontsize{9.5pt}{9.5pt}\selectfont
\begin{aligned}
    \mathcal{A} = 
    & \{ r_{i, j} \; | \; r_i = q \} \; \cup \\
    & \{ r_{i, j} \; | \; q \in T \; \text{and} \; r_i \in T' \; \text{and} \; (T, T') \in \mathcal{L} \} \; \cup \\
    & \{ r_{i, j} \; | \; r_i \in T \; \text{and} \; q \in T' \; \text{and} \; (T, T') \in \mathcal{L} \},
\end{aligned}
\label{eq:attributes}
\fontsize{9.5pt}{9.5pt}\selectfont
\end{equation}
where $r_{i, j}$ is the value of the $j$th attribute column of the $i$th row. Then, based on these attributes (the metadata), we derive their representation $\vq'$ with the encoder: $\vq' = \texttt{Enc}_a(\mathcal{A}; \theta_a)$, described below.

\paragraph{Graph-Structured Set-Encoding}
We now turn to explain how to operationalize the encoding function $\texttt{Enc}_a(\cdot)$, which should effectively represent the diverse attributes $\mathcal{A}$ (over the relational database) into $\vq'$, to enrich the original query representation $\vq$. To accomplish this objective, one possible strategy is to concatenate all the attribute values, and encode the concatenated value with the encoder or append it to the original query (before encoding), following the existing query expansion work~\cite{DBLP:conf/emnlp/ZhengHHH0Y20, Deng2021TowardPA, DBLP:conf/ecir/DholeA24}. However, these approaches have a couple of limitations. First, due to the large volume of data in the database, the number of attributes related to the query could be large, and it might be infeasible to encode their concatenated text with the encoder (due to its limited context length). Also, attributes do not have an inherent order (i.e., permutation invariant), making it arbitrary to determine the sequence in which they should be concatenated.

To tackle these challenges, we propose encoding attributes ($\mathcal{A}$) with a graph-structured set-encoding strategy, which differs from and indeed extends the prior set-encoding~\cite{deep_sets}. Specifically, we first encode every attribute value $r_{i, j}$ in $\mathcal{A}$ into $\vr_{i, j}$ with an attribute encoder: $\vr_{i, j} = \texttt{Enc}_r(r_{i, j}; \theta_r)$, and then aggregate a group of encoded attributes according to each column into the single representation with mean pooling as follows: $\mR_{j} = \texttt{MEAN}(\{ \vr_{i, j} \}_{i=1})$, which then captures the representation of each category (or column) of the metadata. After that, we aggregate all these categorical (column-wise) representations into another representation, which represents the overall metadata for the given query as $\vq' = \texttt{MEAN}(\{ \mR_{j} \}_{j=1})$. Note that this dual-layer structure — aggregating at both the column- and query-levels — resembles a two-layer graph neural network~\cite{GCN, MPNN}, where each layer functionally captures the interactions between the attributes in the same column first and the columns over different tables next in a hierarchical manner.

Let's consider the scenario in Figure~\ref{fig:concept}, where the goal is to retrieve the answer post most likely to be selected as the best by the user. Then, the query is encoded into $\vq$, which is further enriched with the metadata representation $\vq'$ obtained via the proposed graph-structured set-encoding as follows: the metadata ($\mathcal{A}$) includes attributes such as user comments ($\textsc{Comment}$), tags ($\textsc{Tags}$), and the user profile ($\textsc{AboutMe}$); each attribute is encoded into a column-level representation, e.g., $\mR_{\textsc{Comment}} = \texttt{MEAN} ( \{ \texttt{Enc}_r (r_{i, \textsc{Comment}}) \}_{i=1} )$ (and similarly for others); all column-level representations are aggregated into a single query-level representation: $\vq' = \texttt{MEAN} ( \{ \mR_{\textsc{Comment}}, \mR_{\textsc{Tags}}, \mR_{\textsc{AboutMe}} \} )$, which is used to augment the original query representation.

\paragraph{Efficient Training Strategy with Metadata}
\label{sec:method:effi_train}
The number of attributes collected from the relational database is sometimes very large for certain queries, and it may be largely inefficient to consider all of them during training. To address this, we introduce a two-stage sample selection strategy to efficiently train the metadata encoder $\texttt{Enc}_r$ and to efficiently obtain the metadata representation $\vq'$. Specifically, during training, instead of using all attributes in $\mathcal{A}$ for parameter updates, we randomly sample three attributes for each column and use only them to train the metadata encoder. In addition, while we can use all the remaining attributes (without gradients) to obtain the metadata representation along with the representations of three specific attributes for each column (with gradients), using all the remaining attributes may still be time-consuming and may yield the over-fitting issue; thus, we randomly sample some of them and use only them to obtain the representation $\vq'$. Meanwhile, in the inference step, we utilize all the metadata attributes available.

\section{Experimental Setups}
In this section, we describe the main experimental setups. We provide further details in Appendix~\ref{appendix:setup}.

\subsection{Datasets}
Since this is the first work on retrieval that utilizes the relational database for augmenting query representations, we design seven different tasks based on the Stack Exchange and Amazon Product Catalog databases available from~\citet{relbench}, and also the H\&M database from~\citet{relbench/new}.

\paragraph{Stack Exchange}
This dataset is collected from discussions in Stack Exchange\footnote{\url{https://stackexchange.com/}}, an online website for question-answering, and organized into a relational database consisting of seven tables (such as posts, users, and votes). For this dataset, we design two retrieval tasks, as follows:
\noindent \textbf{Answer Retrieval (Any Answer)} involves retrieving any answer posts made by any users in response to a question post. 
\noindent \textbf{Best Answer Retrieval (Best Answer)} is a more challenging task that aims to retrieve a single answer post that has been selected by the owner of the question post. Also, we further consider two different scenarios by dividing the entire dataset by users (\textbf{SplitByUser}) or timestamps (\textbf{SplitByTime}). For the first setting, training, validation, and test sets are divided by users (there are no overlapping users). Similarly, the later setting splits the dataset according to the timestamp that the post was made. For each retrieval instance, the information before the post timestamp is used to augment the query.

\paragraph{Amazon Product Catalog}
This dataset is collected from book reviews on the Amazon Product Catalog, which consists of three tables (users, products, and reviews) over the relational database. For this dataset, we introduce \textbf{Future Purchase Retrieval (Future Purchase)} as the task, which aims to predict any future book purchases of customers based on their current reviews as well as their previous purchases and reviews. Also, we construct two different settings, namely \textbf{ReviewToProduct} and \textbf{ProductToProduct}, where the first one uses the review text as a query while the latter one uses the product description as a query for retrieval.

\paragraph{H\&M}
This dataset includes customer and product data across H\&M's online shopping platforms, consisting of three tables (articles, customers, and transactions). Similar to the Amazon Product Catalog, we consider the \textbf{Future Purchase Retrieval (Future Purchase)} under the \textbf{ProductToProduct} setting, whose goal is to predict future product purchases based on the product history as a query. 

\begin{table*}[t!]
\caption{\small Results on seven different retrieval scenarios using Stack Exchange, Amazon Product Catalog, and H\&M databases.}
\vspace{-0.1in}
\label{tab:main}
\small
\centering
\resizebox{\textwidth}{!}{
\renewcommand{\arraystretch}{1.2}
\setlength{\tabcolsep}{4pt}
\begin{tabular}{cl cc cc cc cc cc cc cc cc}
\toprule

& & \multicolumn{4}{c}{\bf StackExchange (Any Answer)} & \multicolumn{4}{c}{\bf StackExchange (Best Answer)} & \multicolumn{4}{c}{\bf Amazon (Future Purchase)} & \multicolumn{2}{c}{\bf H\&M (Future Purchase)} \\
\cmidrule(l{2pt}r{2pt}){3-6} \cmidrule(l{2pt}r{2pt}){7-10} \cmidrule(l{2pt}r{2pt}){11-14} \cmidrule(l{2pt}r{2pt}){15-16}
& & \multicolumn{2}{c}{\bf SplitByUser} & \multicolumn{2}{c}{\bf SplitByTime} & \multicolumn{2}{c}{\bf SplitByUser} & \multicolumn{2}{c}{\bf SplitByTime} & \multicolumn{2}{c}{\bf ReviewToProduct} & \multicolumn{2}{c}{\bf ProductToProduct} & \multicolumn{2}{c}{\bf ProductToProduct}\\
\cmidrule(l{2pt}r{2pt}){3-4} \cmidrule(l{2pt}r{2pt}){5-6} \cmidrule(l{2pt}r{2pt}){7-8}\cmidrule(l{2pt}r{2pt}){9-10} \cmidrule(l{2pt}r{2pt}){11-12} \cmidrule(l{2pt}r{2pt}){13-14} \cmidrule(l{2pt}r{2pt}){15-16}
 & {\bf Method} & Recall@10 & Acc@100 & Recall@10 & Acc@100 & MRR & Acc@100 & MRR & Acc@100 & Acc@500 & Recall@1000 & Acc@500 & Recall@1000 & Acc@50 & Recall@100 \\
\midrule
& BM25-Anserini & 11.45 & 28.33 & 15.79 & 32.64 & 9.64 & 29.49 & 11.68 & 34.79 & 5.71 & 3.51 & 15.09 & 7.48   & 10.10  & 3.12 \\

\midrule \midrule
\multirowcell{8}[-0.0ex][c]{\rotatebox[origin=c]{90}{\textbf{DPR}}} & No Expan. & 36.15 \scriptsize{± 0.05} & 68.09 \scriptsize{± 0.14} & 35.46 \scriptsize{± 0.55} & 64.48 \scriptsize{± 0.30} & 20.87 \scriptsize{± 0.29} & 56.11 \scriptsize{± 0.09} & 22.87 \scriptsize{± 0.33} & 58.25 \scriptsize{± 0.15} & 6.37 \scriptsize{± 0.49} & 2.74 \scriptsize{± 0.20} & 15.54 \scriptsize{± 0.94} & 7.77 \scriptsize{± 0.24}& 13.80 \scriptsize{± 1.17}  & 5.52 \scriptsize{± 0.66}\\

& Expan. w/ LLM & 32.48 \scriptsize{± 0.26} & 63.79 \scriptsize{± 0.19} & 31.66 \scriptsize{± 0.36} & 60.45 \scriptsize{± 0.43} & 18.37 \scriptsize{± 0.54} & 51.60 \scriptsize{± 0.42} & 20.28 \scriptsize{± 0.32} & 53.61 \scriptsize{± 0.22} &  6.37 \scriptsize{± 0.29} & 2.68 \scriptsize{± 0.10} & 14.32 \scriptsize{± 0.36} &  7.67 \scriptsize{± 0.26}& 13.30 \scriptsize{± 0.29}  & 5.29  \scriptsize{± 0.11} \\

& Expan. w/ LameR &  33.77 \scriptsize{± 0.16} & 65.14 \scriptsize{± 0.30} &  34.09  \scriptsize{± 1.01} &  62.34 \scriptsize{± 0.60} & 20.01  \scriptsize{± 0.38} &  53.24  \scriptsize{± 0.49} & 21.60 \scriptsize{± 0.35} & 55.44  \scriptsize{± 0.76} & 6.31  \scriptsize{± 0.21} & 2.62  \scriptsize{± 0.04} &  15.92 \scriptsize{± 0.57} &  7.87 \scriptsize{± 0.06} & 13.97 \scriptsize{± 0.58}  & 6.05 \scriptsize{± 0.04} \\

& Expan. w/ Query & 36.70 \scriptsize{± 0.30} & 69.15 \scriptsize{± 0.22} & 36.53 \scriptsize{± 0.51} & 66.60 \scriptsize{± 0.38} & 20.48 \scriptsize{± 0.38} & 57.01 \scriptsize{± 0.72} & 22.57 \scriptsize{± 0.23} & 58.94 \scriptsize{± 0.41} & 5.98 \scriptsize{± 0.39} & 2.58 \scriptsize{± 0.11} & 16.61 \scriptsize{± 0.29} &  8.48 \scriptsize{± 0.12}& 13.64 \scriptsize{± 0.87}  & 5.84 \scriptsize{± 0.28} \\

& Expan. w/ User & 36.53 \scriptsize{± 0.06} & 68.26 \scriptsize{± 0.17} & 35.65 \scriptsize{± 0.28} & 65.07 \scriptsize{± 0.15} & 21.66 \scriptsize{± 0.15} &  56.74 \scriptsize{± 0.14} & 23.18 \scriptsize{± 0.06} & 58.81 \scriptsize{± 0.21} & 3.48 \scriptsize{± 0.22} & 2.03 \scriptsize{± 0.10} & 8.75 \scriptsize{± 0.57} & 4.68 \scriptsize{± 0.25}& 13.47 \scriptsize{± 0.58}  & 5.51 \scriptsize{± 0.40} \\

& Expan. w/ Full & 38.76 \scriptsize{± 0.21} & 70.67 \scriptsize{± 0.21} & 38.75 \scriptsize{± 0.48} & 67.37 \scriptsize{± 0.45} & 
20.03 \scriptsize{± 0.38} & 55.00 \scriptsize{± 0.31} & 21.88 \scriptsize{± 0.14} & 56.66 \scriptsize{± 0.33} & 11.04 \scriptsize{± 0.34} & \textbf{6.10} \scriptsize{± 0.24} & 14.67 \scriptsize{± 1.21} & 7.66 \scriptsize{± 0.27}& 6.57 \scriptsize{± 3.50}  & 1.64 \scriptsize{± 0.55}\\

& Expan. w/ Retriever & 38.47 \scriptsize{± 0.34} & 70.37 \scriptsize{± 0.25} & 37.83 \scriptsize{± 0.26} & 66.70 \scriptsize{± 0.15} & 19.54 \scriptsize{± 0.18} & 54.08 \scriptsize{± 0.12} & 21.47 \scriptsize{± 0.26} & 56.14 \scriptsize{± 0.21} & 12.56 \scriptsize{± 0.36} & 5.89 \scriptsize{± 0.25} & 17.29 \scriptsize{± 0.42} & 8.42 \scriptsize{± 0.34}& 9.43 \scriptsize{± 0.58}  & 4.06 \scriptsize{± 0.15}\\

\noalign{\vskip 0.25ex}\cdashline{2-16}\noalign{\vskip 0.75ex}

& DAQu (Ours) & \textbf{41.80} \scriptsize{± 0.27} & \textbf{74.11} \scriptsize{± 0.24} & \textbf{41.67} \scriptsize{± 0.39} & \textbf{71.72} \scriptsize{± 0.33} & \textbf{22.05} \scriptsize{± 0.24} & \textbf{57.81} \scriptsize{± 0.80} &  \textbf{23.70} \scriptsize{± 0.18} & \textbf{59.24} \scriptsize{± 0.46}  & \textbf{13.07} \scriptsize{± 0.19} & 5.97 \scriptsize{± 0.27} & \textbf{17.86} \scriptsize{± 0.39} & \textbf{9.15} \scriptsize{± 0.10}& \textbf{15.49} \scriptsize{± 0.29}  & \textbf{6.63} \scriptsize{± 0.15}\\ 

\midrule \midrule

\multirowcell{8}[-0.0ex][c]{\rotatebox[origin=c]{90}{\textbf{Contriever}}} & No Expan. & 42.08 \scriptsize{± 0.28} & 73.21 \scriptsize{± 0.15} & 41.93 \scriptsize{± 0.07} & 70.08 \scriptsize{± 0.45} & 25.85 \scriptsize{± 0.15} & 64.16 \scriptsize{± 0.34} & 28.37 \scriptsize{± 0.08} & 64.95 \scriptsize{± 0.15} & 8.21 \scriptsize{± 0.32} & 4.63 \scriptsize{± 0.20} & 17.80 \scriptsize{± 0.45} & 9.27 \scriptsize{± 0.06}& 15.15 \scriptsize{± 0.00}  & 5.95 \scriptsize{± 0.00}\\

& Expan. w/ LLM & 38.35 \scriptsize{± 0.63} & 69.35 \scriptsize{± 0.59} & 38.66 \scriptsize{± 0.29} & 66.39 \scriptsize{± 0.20} & 23.27 \scriptsize{± 0.06} & 59.03 \scriptsize{± 0.12} & 25.05 \scriptsize{± 0.33} & 60.32 \scriptsize{± 0.22} & 8.60 \scriptsize{± 0.31} &  4.58 \scriptsize{± 0.20} & 16.82 \scriptsize{± 0.74} & 9.18 \scriptsize{± 0.24}& 15.40 \scriptsize{± 0.36}  & 6.20 \scriptsize{± 0.34} \\

& Expan. w/ LameR & 38.82 \scriptsize{± 0.04} & 69.68 \scriptsize{± 0.02} &   38.78 \scriptsize{± 0.40} &  67.03 \scriptsize{± 0.03} &  24.56 \scriptsize{± 0.22} &  60.12 \scriptsize{± 0.21} & 25.23 \scriptsize{± 0.18} & 59.26 \scriptsize{± 0.46} &  7.26 \scriptsize{± 0.41} &  3.95 \scriptsize{± 0.24} &  16.79 \scriptsize{± 0.46} &  8.73 \scriptsize{± 0.04} & 15.15 \scriptsize{± 0.00}  & 5.91 \scriptsize{± 0.08} \\

& Expan. w/ Query & 41.84 \scriptsize{± 0.31} & 73.96 \scriptsize{± 0.11} & 42.92 \scriptsize{± 0.13} & 71.54 \scriptsize{± 0.45} & 24.11 \scriptsize{± 0.53} &  63.39 \scriptsize{± 0.35} & 27.67 \scriptsize{± 0.11} & 65.03 \scriptsize{± 0.40} & 8.93 \scriptsize{± 0.36} & 4.68 \scriptsize{± 0.17} & 18.13 \scriptsize{± 0.58} & 9.31 \scriptsize{± 0.07}& \textbf{15.66} \scriptsize{± 0.00}  & 6.02 \scriptsize{± 0.06} \\

& Expan. w/ User & 42.21 \scriptsize{± 0.36} & 73.45 \scriptsize{± 0.21} & 42.26 \scriptsize{± 0.41} & 70.22 \scriptsize{± 0.20} & 25.93 \scriptsize{± 0.15} &  62.87 \scriptsize{± 0.25} & 28.20 \scriptsize{± 0.12} & 64.67 \scriptsize{± 0.26} & 6.34 \scriptsize{± 0.26} & 2.55 \scriptsize{± 0.15} & 7.23  \scriptsize{± 0.54} & 4.35 \scriptsize{± 0.44}& 11.28 \scriptsize{± 0.29}  & 4.70 \scriptsize{± 0.39} \\

&  Expan. w/ Full & 45.25 \scriptsize{± 0.24} & 76.20 \scriptsize{± 0.17} & 44.43 \scriptsize{± 0.13} & 72.50 \scriptsize{± 0.18} & 26.01 \scriptsize{± 0.27} & 63.59 \scriptsize{± 0.23} & 28.21 \scriptsize{± 0.10} & 64.06 \scriptsize{± 0.36} & 17.23 \scriptsize{± 0.46} & 8.86 \scriptsize{± 0.22} & 17.02 \scriptsize{± 0.89} & 9.37 \scriptsize{± 0.53}& 5.39 \scriptsize{± 0.29}  & 1.92 \scriptsize{± 0.30}\\

& Expan. w/ Retriever & 44.69 \scriptsize{± 0.25} & 75.52 \scriptsize{± 0.23} & 44.66 \scriptsize{± 0.27} & 72.24 \scriptsize{± 0.39} & 24.71 \scriptsize{± 0.18} & 62.15 \scriptsize{± 0.24} & 27.28 \scriptsize{± 0.25} & 63.52 \scriptsize{± 0.55} & 17.71 \scriptsize{± 0.22} & 7.18 \scriptsize{± 0.55} & 17.71 \scriptsize{± 0.22} & 9.40 \scriptsize{± 0.21}& 13.13 \scriptsize{± 0.87}  & 4.99 \scriptsize{± 0.05}\\

\noalign{\vskip 0.25ex}\cdashline{2-16}\noalign{\vskip 0.75ex}
& DAQu (Ours) & \textbf{49.74} \scriptsize{± 0.26} & \textbf{80.27} \scriptsize{± 0.23} & \textbf{50.28} \scriptsize{± 0.49} & \textbf{78.06} \scriptsize{± 0.38} & \textbf{26.47} \scriptsize{± 0.26} & \textbf{65.16} \scriptsize{± 0.33} & \textbf{28.82} \scriptsize{± 0.07} & \textbf{65.47} \scriptsize{± 0.58} & \textbf{18.75} \scriptsize{± 0.91} & \textbf{9.86} \scriptsize{± 0.46} & \textbf{19.87} \scriptsize{± 0.44} & \textbf{10.42} \scriptsize{± 0.67}& 15.40 \scriptsize{± 0.36}  & \textbf{6.25} \scriptsize{± 0.34}\\

\midrule \midrule

\multirowcell{8}[-0.0ex][c]{\rotatebox[origin=c]{90}{\textbf{BGE-M3}}} 

& No Expan. & 39.83 \scriptsize{± 0.33} & 71.08 \scriptsize{± 0.06} & 39.54 \scriptsize{± 0.44} & 68.02 \scriptsize{± 0.27} & 22.37\scriptsize{± 0.23} & 58.41 \scriptsize{± 0.39} &  22.96 \scriptsize{± 0.20} & 57.24 \scriptsize{± 0.73} & 7.59 \scriptsize{± 0.15} & 3.87 \scriptsize{± 0.03} & 16.10 \scriptsize{± 0.05} & 8.29 \scriptsize{± 0.18}&  14.65 \scriptsize{± 0.00}  & 5.59  \scriptsize{± 0.17} \\

& Expan. w/ LLM &  37.57 \scriptsize{± 0.20} & 67.24 \scriptsize{± 0.47} & 37.52 \scriptsize{± 0.37} & 64.29 \scriptsize{± 0.20} & 19.21 \scriptsize{± 0.13} & 51.52 \scriptsize{± 0.66} & 19.95\scriptsize{± 0.18} & 51.72 \scriptsize{± 0.28} & 8.27 \scriptsize{± 1.60} & 3.75 \scriptsize{± 0.40} & 15.98 \scriptsize{± 0.31} & 8.00 \scriptsize{± 0.09}& 14.81 \scriptsize{± 0.29}  &  6.05 \scriptsize{± 0.15} \\

& Expan. w/ LameR & 38.46 \scriptsize{± 0.34} & 68.07 \scriptsize{± 0.13} &  38.06 \scriptsize{± 0.34} & 65.37  \scriptsize{± 0.19} &  20.42 \scriptsize{± 0.46} &  52.41 \scriptsize{± 0.19} &  21.07 \scriptsize{± 0.16} & 53.51  \scriptsize{± 0.62} &  7.32 \scriptsize{± 0.77} & 3.67  \scriptsize{± 0.66} & 15.48  \scriptsize{± 0.57} &  8.20 \scriptsize{± 0.17} & 14.14 \scriptsize{± 0.00}  & 5.82 \scriptsize{± 0.25} \\

& Expan. w/ Query & 39.90 \scriptsize{± 1.16} & 72.15 \scriptsize{± 0.31} &  40.64 \scriptsize{± 0.68} & 70.09 \scriptsize{± 0.26} & 22.96 \scriptsize{± 0.57} & 60.32 \scriptsize{± 0.79} & 23.07 \scriptsize{± 0.50} & 58.95 \scriptsize{± 0.84} & 7.41 \scriptsize{± 0.46} & 3.75 \scriptsize{± 0.36} & 16.16 \scriptsize{± 0.31} & 8.25 \scriptsize{± 0.07}& 15.32 \scriptsize{± 0.29}  & 6.08 \scriptsize{± 0.07} \\

& Expan. w/ User &  42.10 \scriptsize{± 0.46} & 73.13 \scriptsize{± 0.18} & 41.60 \scriptsize{± 0.23} & 69.82 \scriptsize{± 0.04} & 22.84 \scriptsize{± 0.80} & 59.74 \scriptsize{± 0.93} & 23.43 \scriptsize{± 0.19} & 58.47 \scriptsize{± 0.07} & 4.49 \scriptsize{± 1.19} & 1.91 \scriptsize{± 0.05} & 11.79 \scriptsize{± 0.31} & 5.01 \scriptsize{± 0.27}& 15.15 \scriptsize{± 0.00}  & 6.01 \scriptsize{± 0.36} \\

&  Expan. w/ Full &  41.47 \scriptsize{± 0.19} & 73.00 \scriptsize{± 0.10} & 41.63 \scriptsize{± 0.90} & 70.06 \scriptsize{± 0.60} & 23.42 \scriptsize{± 0.17} & 58.11 \scriptsize{± 1.06} & 23.17 \scriptsize{± 0.09} & 57.29 \scriptsize{± 0.08} & 13.10 \scriptsize{± 0.05} & \textbf{7.36} \scriptsize{± 0.47} & 15.03 \scriptsize{± 1.60} & 8.12 \scriptsize{± 1.87}& 4.88 \scriptsize{± 0.29}  & 1.68 \scriptsize{± 0.36} \\

& Expan. w/ Retriever & 41.77 \scriptsize{± 0.46} & 72.76 \scriptsize{± 0.24} & 41.79 \scriptsize{± 0.23} & 70.00 \scriptsize{± 0.23} & 22.84 \scriptsize{± 0.21} & 58.36 \scriptsize{± 0.36} & 22.44 \scriptsize{± 0.42} & 56.25 \scriptsize{± 0.67} & 12.92 \scriptsize{± 0.26} & 6.13 \scriptsize{± 0.15} & 17.56 \scriptsize{± 0.57} & 8.56 \scriptsize{± 0.17}& 13.30 \scriptsize{± 1.17}  & 5.49 \scriptsize{± 0.14} \\

\noalign{\vskip 0.25ex}\cdashline{2-16}\noalign{\vskip 0.75ex}

& DAQu (Ours) & \textbf{44.92} \scriptsize{± 0.22} & \textbf{75.67} \scriptsize{± 0.05} &  \textbf{45.26} \scriptsize{± 0.39} & \textbf{73.61} \scriptsize{± 0.07} & \textbf{24.47} \scriptsize{± 0.45} & \textbf{61.55} \scriptsize{± 0.18} & \textbf{24.20} \scriptsize{± 0.01} & \textbf{59.26} \scriptsize{± 0.24} & \textbf{14.67} \scriptsize{± 0.88} & 6.93 \scriptsize{± 0.85} & \textbf{18.21} \scriptsize{± 0.15} & \textbf{9.03} \scriptsize{± 0.33}& \textbf{15.66} \scriptsize{± 0.00}  & \textbf{6.86} \scriptsize{± 0.05} \\

\bottomrule

\end{tabular}
}
\vspace{-0.05in}
\end{table*}

\subsection{Models}
We explain the backbone retrieval models and the query augmentation baselines that we compare.

\paragraph{Retrieval Models}
We use three dense retrievers: \textbf{DPR} is a dense retrieval model trained with a pair of a query and its relevant document~\cite{dpr};
\textbf{Contriever} is another dense retriever, but is trained in an unsupervised fashion~\cite{contriever};
\textbf{BGE-M3} is a recent dense retriever designed to enhance generalization across diverse retrieval tasks~\cite{bge-m3}. As an indicator, we report the results of the sparse retriever (\textbf{BM25}).

\paragraph{Augmentation Models}
We compare our DAQu against relevant query augmentation baselines: 
\noindent \textbf{1) No Expansion (No Expan.):} 
This model uses the query for retrieval without expansion.
\noindent \textbf{2) Query Expansion w/ LLM (Expan. w/ LLM):}
This model utilizes the capability of LLMs, prompting them to generate query-related pseudo-documents that are expanded to queries~\cite{DBLP:conf/emnlp/WangYW23}.
\noindent \textbf{3) Query Expansion w/ LameR (Expan. w/ LameR):}
This model similarly utilizes LLMs but further augments them with query-relevant documents via retrieval for query expansion~\cite{lamer}. 
\noindent \textbf{4) Query Expansion w/ Query associated Table (Expan. w/ Query):}
This model expands queries with the information sourced from the query-related single data store (table), following~\citet{DBLP:conf/sigir/ZhangDL20}. 
\noindent \textbf{5) Query Expansion w/ User associated Table (Expan. w/ User):}
Similarly, this model expands queries with the user-related table, following~\citet{DBLP:conf/vldb/BussMTT0L23}.
\noindent \textbf{6) Full Metadata Expansion (Expan. w/ Full):}
This model concatenates queries with all textual terms of the associated metadata from the database (spanning multiple tables).
\noindent \textbf{7) Query Expansion w/ Retriever (Expan. w/ Retriever):}
Similar to~\citet{Deng2021TowardPA}, this model also appends the metadata terms to the queries. Yet, before expansion, it employs a retriever (BM25) to select terms that are most relevant to the query, and only these selected terms are appended.
\noindent \textbf{8) DAQu (Ours):}
This is our model that augments the query representation by incorporating the metadata representation on a latent space, obtained by graph-structured set-encoding.

\subsection{Evaluation Metrics}
We use the following metrics:
\noindent \textbf{1) Accuracy@K (Acc@K)} determines the fraction of queries for which the top-$k$ results include at least one relevant document.
\noindent \textbf{2) Recall@K} calculates the percentage of all relevant documents that are present within the top-$k$ results.
\noindent \textbf{3) Mean Reciprocal Rank (MRR)} computes the average of the inverse of the ranks at which the first relevant document is found across queries.
\noindent \textbf{4) Mean Average Precision (MAP)} measures the mean precision score calculated after each relevant document is retrieved, across all queries.

\subsection{Implementation Details}
We train all retrieval models with a learning rate of 2e-5 with AdamW~\cite{adamw}. Also, we set $\lambda$ as 0.7 chosen based on a search within the range of \{0.1, 0.3, 0.5, 0.7, 0.9\}, and randomly sample 30 features for the no-gradient metadata features in our efficient training strategy and 3 features for gradient updates. Regarding metrics, for answer post retrieval on Stack Exchange, which aligns more closely with conventional document retrieval tasks, we use a diverse range of K values, including 10, 20, 50, and 100. In contrast, for product retrieval with Amazon Product Catalog, where the goal is not only to identify items of interest but specifically those the user will purchase, considering the long-tail nature of product recommendations, we use larger K values of 500 and 1000, following prior work on product retrieval \cite{DBLP:conf/kdd/LiLJLYZWM21, DBLP:conf/www/WangLZZWXLY23, DBLP:journals/corr/abs-2407-19829}. Lastly, we report the average of three different runs.

\section{Experimental Results and Analyses}
We now present the results and detailed analyses.

\begin{figure*}[t!]
    \begin{minipage}{0.30\linewidth}
        \centering
        \includegraphics[width=0.91\columnwidth]{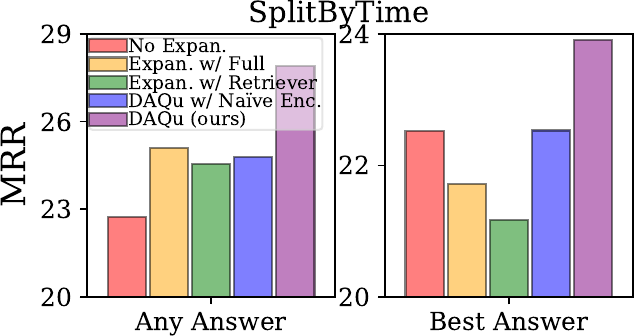}
        \label{fig:mean_pooling}
        \vspace{-0.1in}
        \caption{\small Results of the set encoding strategy of DAQu over naïve encoding, simply aggregating all representations.}
        \label{fig:meanpooling}
    \end{minipage}
    \hfill
    \begin{minipage}{0.34\linewidth}
        \begin{minipage}{0.49\linewidth}
            \centering
            \includegraphics[width=0.95\columnwidth]{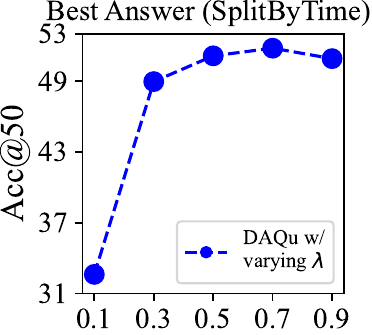}
            \label{fig:lambda}
        \end{minipage}
        \begin{minipage}{0.49\linewidth}
            \centering
            \includegraphics[width=0.95\columnwidth]{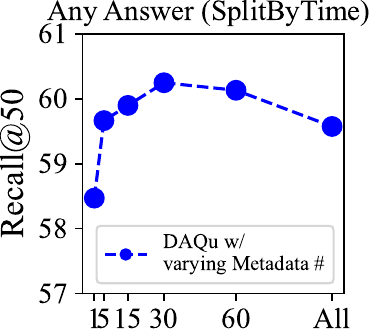}
            \label{fig:metanum_train_line}
        \end{minipage}
    \vspace{-0.1in}
    \caption{\small Results by varying lambda values (Left) and the number of metadata features within each category for training (Right).}
    \label{fig:minipage}
    \end{minipage}
    \hfill
    \begin{minipage}{0.33\textwidth}
        \centering
        \captionof{table}{\small Reranking results following retrieval with DAQu on the SplitByUser scenario of StackExchange (Any Answer).}
        \label{tab:rank}
        \vspace{-0.1in}
        \renewcommand{\arraystretch}{0.99}
        \setlength{\tabcolsep}{6pt}
        \resizebox{\linewidth}{!}{%
            \begin{tabular}{llcccc}
            \toprule
            & \textbf{Methods} & \textbf{Recall@5} & \textbf{Recall@10} & \textbf{Acc@5} & \textbf{Acc@10} \\
            \midrule
            \midrule
            \multirow{4}{*}{\rotatebox[origin=c]{90}{\scriptsize{
             \textbf{Contriever}}}}
            
             & \textbf{No Expan.} & 28.95 & 36.79 & 51.00 &  59.67 \\
             & \textbf{No Expan. + Rerank} & 35.78 & 43.09 & 60.33 & 67.67 \\
            
            \cdashline{2-6}\noalign{\vskip 0.75ex}
            
             & \textbf{DAQu (Ours)} & 35.53 & 43.78 & 59.33 & 68.33 \\
             & \textbf{DAQu (Ours) + Rerank} & \textbf{42.62} & \textbf{49.69} & \textbf{69.33} & \textbf{75.67} \\
             
            \bottomrule
            \toprule
            
            \multirow{4}{*}{\rotatebox[origin=c]{90}{\scriptsize{
             \textbf{BGE-M3}}}}
            
             & \textbf{No Expan.} & 27.20 & 34.39 & 50.33 & 58.67 \\
             & \textbf{No Expan. + Rerank} & 34.40 & 38.99 & 58.33 & 62.33 \\
            
            \cdashline{2-6}\noalign{\vskip 0.75ex}
            
             & \textbf{DAQu (Ours)} & 32.89 & 40.40 & 54.67 & 63.33  \\
             & \textbf{DAQu (Ours) + Rerank} & \textbf{40.19} & \textbf{45.64} & \textbf{62.33} & \textbf{66.33} \\
             
            \bottomrule
            \end{tabular}
        }
    \end{minipage}
    \vspace{-0.075in}
\end{figure*}

\paragraph{Main Results}
We report the overall results across seven different tasks with multiple databases in Table~\ref{tab:main}. From this, we find that DAQu outperforms all baselines substantially, demonstrating the effectiveness of our approach that augments queries with their corresponding metadata representations (obtained from graph-based set-encoding). We provide the results with additional metrics in Appendix~\ref{sup:additional_result}.

To be specific, our findings reveal that expanding queries with LLMs themselves is suboptimal as their parametric knowledge lacks information specific to each user and its query, which relies instead on general patterns stored within them. In contrast, expanding queries with information from a single source of external data stores (Expan. w/ Query and Expan. w/ User) achieves decent performance improvements over the no-expansion baseline, highlighting the importance of incorporating query-specific and user-specific information during query augmentation. Furthermore, leveraging multiple relational tables from the database, such as Expan. w/ Full and Expan. w/ Retriever, further enhances retrieval performances, which underscores the value of considering interrelated information over the relational database for query expansion.

Notably, the proposed DAQu demonstrates substantial improvements across all tasks over all baselines, highlighting the effectiveness of our proposed set-encoding strategy for incorporating metadata into query representations. For example, in the Answer Retrieval task with Stack Exchange, DAQu achieves performance improvements of 18.73\% and 16.91\% on SplitByUser and SplitByTime settings, respectively, in Recall@10. Also, DAQu consistently shows superior performance on the Best Answer Retrieval task, which is more complicated (since the model should retrieve the single post that the user would select as the best one, requiring both the query-specific and user-specific information), where diverse expansion models even degrade the performance over the baseline without expansion. Finally, the superior performance of DAQu on the Future Purchase Retrieval task further confirms that it can be applicable to diverse retrieval tasks. 

\paragraph{Effectiveness of Set-Encoding}
To see the effectiveness of the graph-based set-encoding strategy when incorporating the metadata information into the query, we compare it with two types of baselines: appending their textual terms into the query or encoding them without considering the graph structure. As Figure~\ref{fig:meanpooling} shows, simply appending the query with additional terms or taking the average of all representations in the metadata without graph structure is not as effective as ours. This demonstrates the efficacy of our two-stage (column- and query-levels) set-based metadata encoding strategy.

\paragraph{Analyses on Hyperparameters}
We explore how varying the lambda value ($\lambda$) (balancing the query and metadata representations) impacts the overall results in Figure~\ref{fig:minipage}.
Specifically, when the lambda value is too low ($\lambda$ = 0.1), the model fails to capture the original query's intent. Conversely, a high lambda value ($\lambda = 0.9$) leads to the model overemphasizing the original query over the metadata, thereby underutilizing the meaningful metadata representation, which degrades the performance.
Thus, selecting an optimal lambda value is crucial for balancing these aspects to enhance performance.

We further investigate the impact of varying the number of no-gradient metadata features for each category on overall performance when training the DAQu model. Figure~\ref{fig:minipage} shows that a low count of metadata features per category results in reduced performance, indicating the importance of sufficient features for enhanced results. However, using all metadata features is not only inefficient but also degrades performance. Therefore, it is essential to select the appropriate number of metadata features to optimize model efficiency and effectiveness.

\paragraph{Effectiveness of Reranking with DAQu}
To see whether DAQu provides high-quality candidate sets for reranking, we also conduct an auxiliary analysis, applying the strong reranker~\cite{rankgpt} on top of the retrieval results from DAQu. As shown in Table~\ref{tab:rank}, reranking leads to substantial performance improvements across all models (while it introduces a slight efficiency trade-off), with DAQu combined with reranking achieving the best results.

\begin{table}[t!]
\caption{\small Ablation studies involving the removal or addition of each metadata category on Any Answer (SplitByTime), where Q. and A. refer to question and answer posts, respectively.}
\vspace{-0.1in}
\label{tab:ablation}
\small
\centering
\resizebox{0.475\textwidth}{!}{
\renewcommand{\arraystretch}{1.0}
\renewcommand{\tabcolsep}{2.5mm}
\begin{tabular}{lccccc}
\toprule
& \multicolumn{2}{c}{\textbf{Recall}} & \multicolumn{2}{c}{\textbf{Accuracy}} \\
\cmidrule(lr){2-3} \cmidrule(lr){4-5}
\textbf{Metadata Category} & \textbf{R@20} & \textbf{Increase.} & \textbf{Acc@20} & \textbf{Increase.} \\

\midrule
\midrule

\multirowcell{1}[-0.0ex][l]{\textbf{DAQu (Ours)}}
& 49.93 &  &  54.44 &  \\

\noalign{\vskip 0.25ex}\cdashline{1-5}\noalign{\vskip 0.75ex}

\multirowcell{1}[-0.0ex][l]{\textbf{w/o Comments in Q.}}
& 46.75 & -6.38\% & 51.14  & -6.06\%   \\

\multirowcell{1}[-0.0ex][l]{\textbf{w/o Comments in A.}} 
& 46.06 & -7.74\% & 50.57  & -7.11\%  \\

\multirowcell{1}[-0.0ex][l]{\textbf{w/o Tags in Q.}} 
& 49.61 & -0.63\% & 54.29  & -0.28\%  \\

\noalign{\vskip 0.25ex}\cdashline{1-5}\noalign{\vskip 0.75ex}

\multirowcell{1}[-0.0ex][l]{\textbf{No Expan. }}
& 42.22 &  &  46.39 &   \\

\noalign{\vskip 0.25ex}\cdashline{1-5}\noalign{\vskip 0.75ex}

\multirowcell{1}[-0.0ex][l]{\textbf{w/ Comments in Q.}}
& 45.24 & +7.14\% & 49.69  &  +7.10\% \\

\multirowcell{1}[-0.0ex][l]{\textbf{w/ Comments in A.}} 
& 47.89 & +13.41\% & 52.31  & +12.76\%  \\

\multirowcell{1}[-0.0ex][l]{\textbf{w/ Tags in Q.}} 
& 43.60 & +3.27\% & 47.93  &  +3.31\% \\
\bottomrule
\end{tabular}
}
\vspace{-0.075in}
\end{table}

\paragraph{Analyses on Metadata Category}
To investigate how each category of the metadata contributes to overall performance, we conduct ablation studies, reporting the rate of performance increase when excluding or adding each category, with DPR.
As Table~\ref{tab:ablation} shows, each category plays a crucial role in performance gains.
Also, while each category contributes to improved performance compared to the baseline without expansion, their performances are still not as high as when all categories are used, which implies that the information from each category is complementary to each other.
Interestingly, using the `tags' category (the information within the same table as the query) provides a small improvement, compared to using the `comments' category from another table, which corroborates our hypothesis that it is important to use knowledge from multiple tables over the relational database.

\begin{table}[t!]
\caption{\small Results on efficiency, based on elapsed and relative time per query, by varying the number of metadata features for category during inference on Any Answer (SplitByTime).}
\vspace{-0.1in}
\label{tab:metadata_inference}
\small
\centering
\resizebox{0.48\textwidth}{!}{
\renewcommand{\arraystretch}{1.0}
\renewcommand{\tabcolsep}{2.5mm}
\begin{tabular}{lccccc}
\toprule
& \multicolumn{2}{c}{\textbf{Efficiency}} & \multicolumn{2}{c}{\textbf{Effectiveness}} \\
\cmidrule(lr){2-3} \cmidrule(lr){4-5}
\textbf{\# of Metadata} & \textbf{Elpased} & \textbf{Relative} & \textbf{MAP} & \textbf{Acc@100} \\

\midrule
\midrule

\multirowcell{1}[-0.0ex][l]{\textbf{No Expan.}}
&  0.062 & 1 & 22.94 & 64.15 \\
\multirowcell{1}[-0.0ex][l]{\textbf{Expan. w/ Full}}
&  0.062 & 1.002 & 25.09 & 67.31 \\

\noalign{\vskip 0.25ex}\cdashline{1-5}\noalign{\vskip 0.75ex}

\multirowcell{1}[-0.0ex][l]{\textbf{1 per Category}} 
& 0.073  & 1.182 & 24.06 & 67.99 \\

\multirowcell{1}[-0.0ex][l]{\textbf{2 per Category}} 
& 0.074  &  1.20 & 26.69 & 70.64 \\

\multirowcell{1}[-0.0ex][l]{\textbf{3 per Category}} 
& 0.074  &  1.205 & 27.30 & 71.57 \\

\multirowcell{1}[-0.0ex][l]{\textbf{All per Category}} 
&  0.075 & 1.218 & 27.53 & 71.98 \\

\bottomrule
\end{tabular}
}
\vspace{-0.075in}
\end{table}

\paragraph{Analyses on Inference Efficiency}
We extend our investigation to the efficiency in inference, by varying the number of metadata features used for query augmentation. As Table~\ref{tab:metadata_inference} shows, although using all the metadata features during inference is effective, it requires more time compared to the model without expansion. By contrast, employing a small number of metadata features enhances efficiency while sacrificing performance. The results indicate that, at a certain point (e.g., 3 features per category), there is a region where we can achieve reasonable performance alongside improved efficiency.

\paragraph{Case Study}
Lastly, we provide qualitative case studies of our DAQu and its error analysis in Appendix~\ref{appendix:case} and Appendix~\ref{appendix:error}, respectively.

\section{Conclusion}
In this work, we presented a novel query augmentation framework, DAQu, which enhances the representation of the query with its relevant information within multiple tables over the database. To utilize the metadata features at scale with order invariance, we proposed graph-based set-encoding, which hierarchically aggregates column-level and query-level information. We validated our DAQu on seven different retrieval tasks designed with various databases, showcasing the effectiveness of our database-augmented query representation.

\section*{Limitations}
While our DAQu framework effectively represents the diverse pieces of query-related metadata (over the relational database) through a graph-structured set-encoding strategy, the process of encoding and aggregating metadata representations at both the column- and query-levels may pose efficiency challenges in real-world applications. To address these concerns, we conducted a detailed analysis of the trade-off between the effectiveness and efficiency of DAQu in Table~\ref{tab:metadata_inference}, and showcased that our approach can significantly enhance the effectiveness only with a marginal compensation of the efficiency. On the other hand, this finding still suggests that investigating more advanced methods to further increase run-time efficiency (such as data pruning) would be a valuable direction for future work. Furthermore, the database-augmented retrieval tasks that we designed seem to be quite challenging for the retrieval models. While DAQu generally shows significantly improved performance, there is still a large room for further improving retrieval performance (which we slightly addressed by introducing the reranker in Table~\ref{tab:rank}). Lastly, we wanted to make sure that our framework is validated in realistic retrieval scenarios with real-world large-scale relational databases; however, many such databases are commonly used in enterprise settings and are rarely made publicly available, making it challenging to establish such experimental benchmarking setups. While we validated ours on recently released, real-world relational databases from Stack Exchange, Amazon, and H\&M, developing and releasing more databases would be of interest to the community.

We then would like to discuss interesting avenues for future work, which are orthogonal to the focus of our work and lie far beyond its scope. First, while our focus is on retrieval, a promising avenue is to extend DAQu to downstream applications, such as Retrieval-Augmented Generation (RAG)~\cite{DBLP:journals/debu/ChristmannW24,HybGRAG}, by leveraging the fine-grained and up-to-date user and content information stored in relational databases.Also, it would be valuable to explore a broader challenge faced by many retrieval systems: the trade-off between relevance and exploration. Our work primarily focuses on improving retrieval relevance by leveraging query-associated metadata, as reflected in the performance improvements in Table~\ref{tab:main}. Also, Appendix~\ref{appendix:diversity} further indicates minimal risk of overly narrow results due to metadata. However, in real-world applications, retrieval systems often need to balance relevance with exploration, surfacing diverse or novel content beyond users’ historical interests. This challenge, though important, falls outside the scope of our work, as it requires different research assumptions and techniques (such as counterfactual user modeling or diversity-oriented prompting) that are orthogonal to our metadata-driven augmentation framework, leaving them for future investigation.

\section*{Ethics Statement}
A retrieval system can enhance the factual grounding of recent LLMs when it is integrated with them, which helps prevent the generation of plausible but incorrect answers. We believe that, following this line of directions, our DAQu can play a crucial role in diverse retrieval-augmented generation applications. Yet, it is important to note that as relational databases contain substantial amounts of knowledge, including personal information, some potential privacy concerns must be carefully managed when utilizing this information. In other words, further development of filtering strategies that tag and mask personal information across multiple tables before delivery to users or integration with LLMs would be required for real-world applications.

\section*{Acknowledgements}
This work was supported by Institute for Information and communications Technology Promotion (IITP) grant funded by the Korea government (No. 2018-0-00582, Prediction and augmentation of the credibility distribution via linguistic analysis and automated evidence document collection).

\bibliography{custom}

\appendix

\clearpage

\appendix

\section{Implementation Details}
\label{appendix:setup}

\subsection{Datasets}

\begin{table}[t!]
\caption{Data statistics for each task designed with StackExchange, Amazon Product Catalog, and H\&M databases.}
\vspace{-0.1in}
\label{tab:data_stats}
\small
\centering
\resizebox{\columnwidth}{!}{
\renewcommand{\arraystretch}{1.0}
\begin{tabular}{ll ccc}
\toprule
\textbf{Task} & \textbf{Setting} & \textbf{Training} & \textbf{Valid} & \textbf{Test} \\ \midrule
\multicolumn{5}{c}{\textit{StackExchange}} \\ \midrule
\multirowcell{2}[-0.0ex][l]{Any Answer} & SplitByUser & 128,981 & 17,132 & 15,583 \\ 
 & SplitByTime & 130,398 & 15,861 & 15,437 \\
 \noalign{\vskip 0.25ex}\cdashline{1-5}\noalign{\vskip 0.75ex}
\multirowcell{2}[-0.0ex][l]{Best Answer} & SplitByUser & 43,889 & 6,106 & 5,252 \\
& SplitByTime & 42,900 & 6,018 & 6,329 \\ \midrule
\multicolumn{5}{c}{\textit{Amazon Product Catalog}} \\ \midrule
\multirowcell{2}[-0.0ex][l]{Future Purchase} & ReviewToProduct & \multirowcell{2}[-0.0ex][c]{65,797} & \multirowcell{2}[-0.0ex][c]{4,561} & \multirowcell{2}[-0.0ex][c]{5,956} \\
& ProductToProduct & & & \\ \midrule
\multicolumn{5}{c}{\textit{H\&M}} \\ \midrule
\multirowcell{1}[-0.0ex][l]{Future Purchase} & ProductToProduct & \multirowcell{1}[-0.0ex][c]{24,479} & \multirowcell{1}[-0.0ex][c]{1,133} & \multirowcell{1}[-0.0ex][c]{1,124} \\
\bottomrule

\end{tabular}
}
\vspace{-0.1in}
\end{table}

\begin{figure*}
    \centering
    \includegraphics[width=0.975\linewidth]{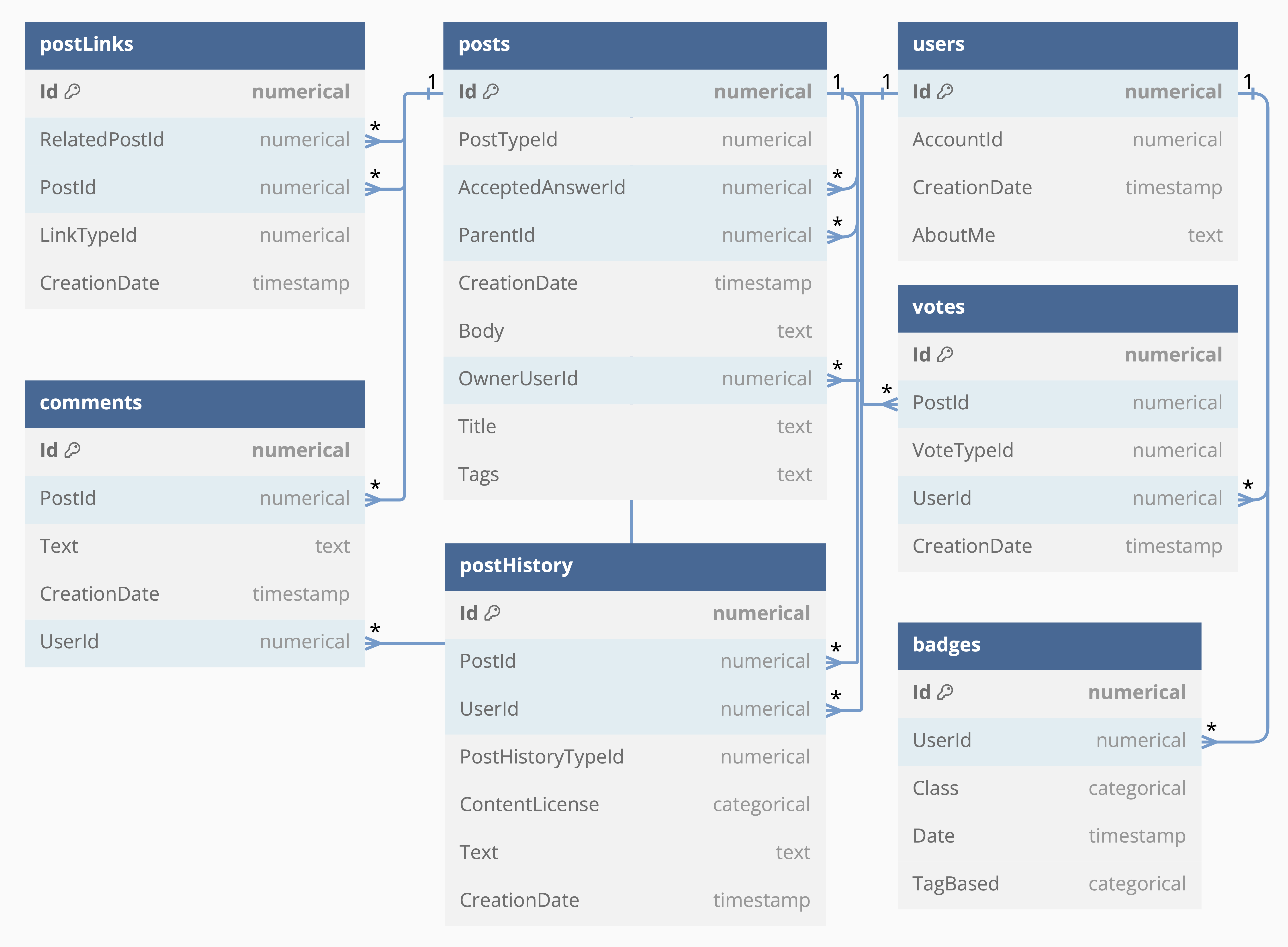}
    \vspace{-0.05in}
    \caption{A database schema for Stack Exchange, which is provided from~\citet{relbench}.}
    \label{fig:schema:stack}
\end{figure*}

\begin{figure*}
    \centering
    \includegraphics[width=0.975\linewidth]{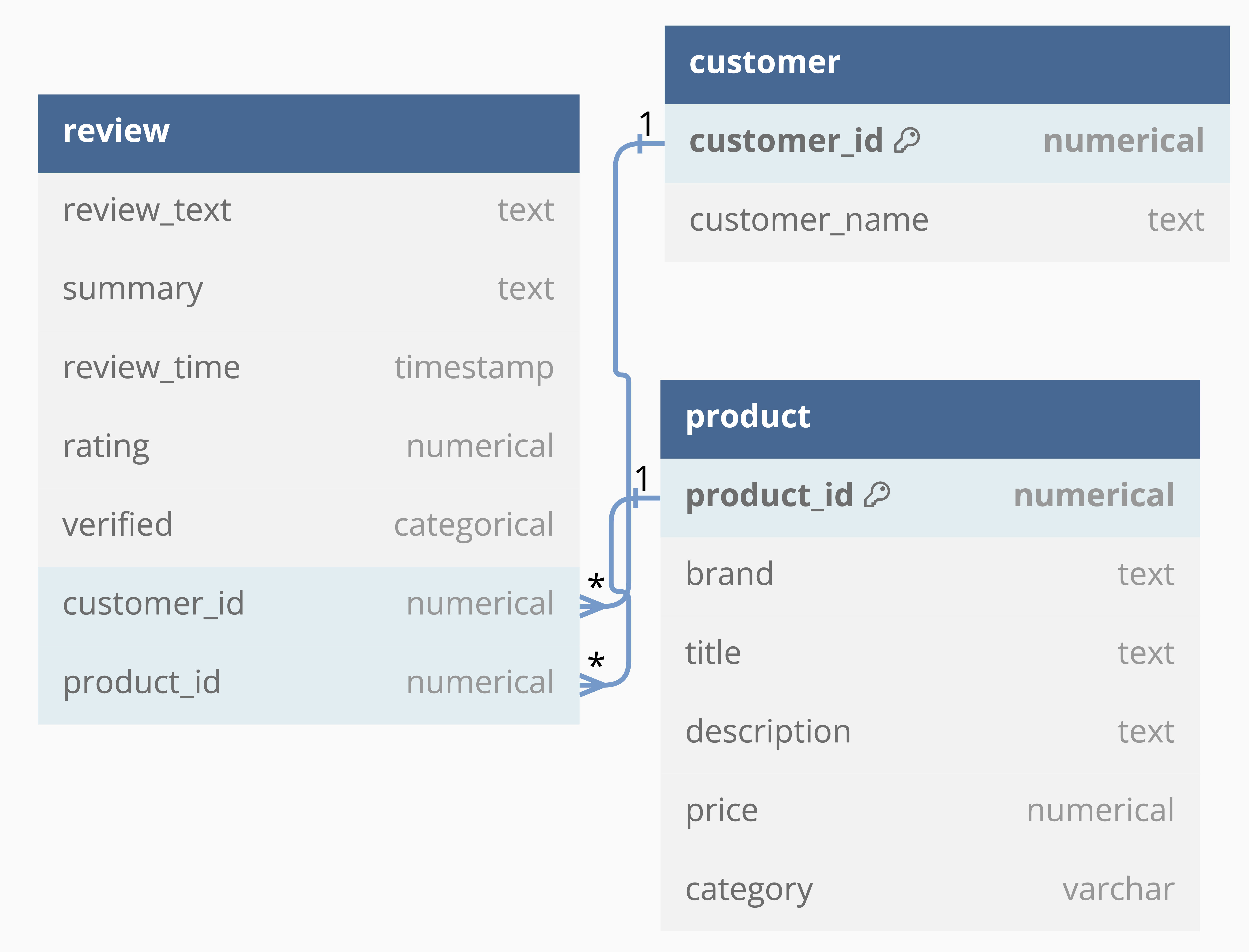}
    \vspace{-0.05in}
    \caption{A database schema for Amazon Product Catalog, which is provided from~\citet{relbench}.}
    \label{fig:schema:amazon}
\end{figure*}

\begin{figure*}
    \centering
    \includegraphics[width=0.975\linewidth]{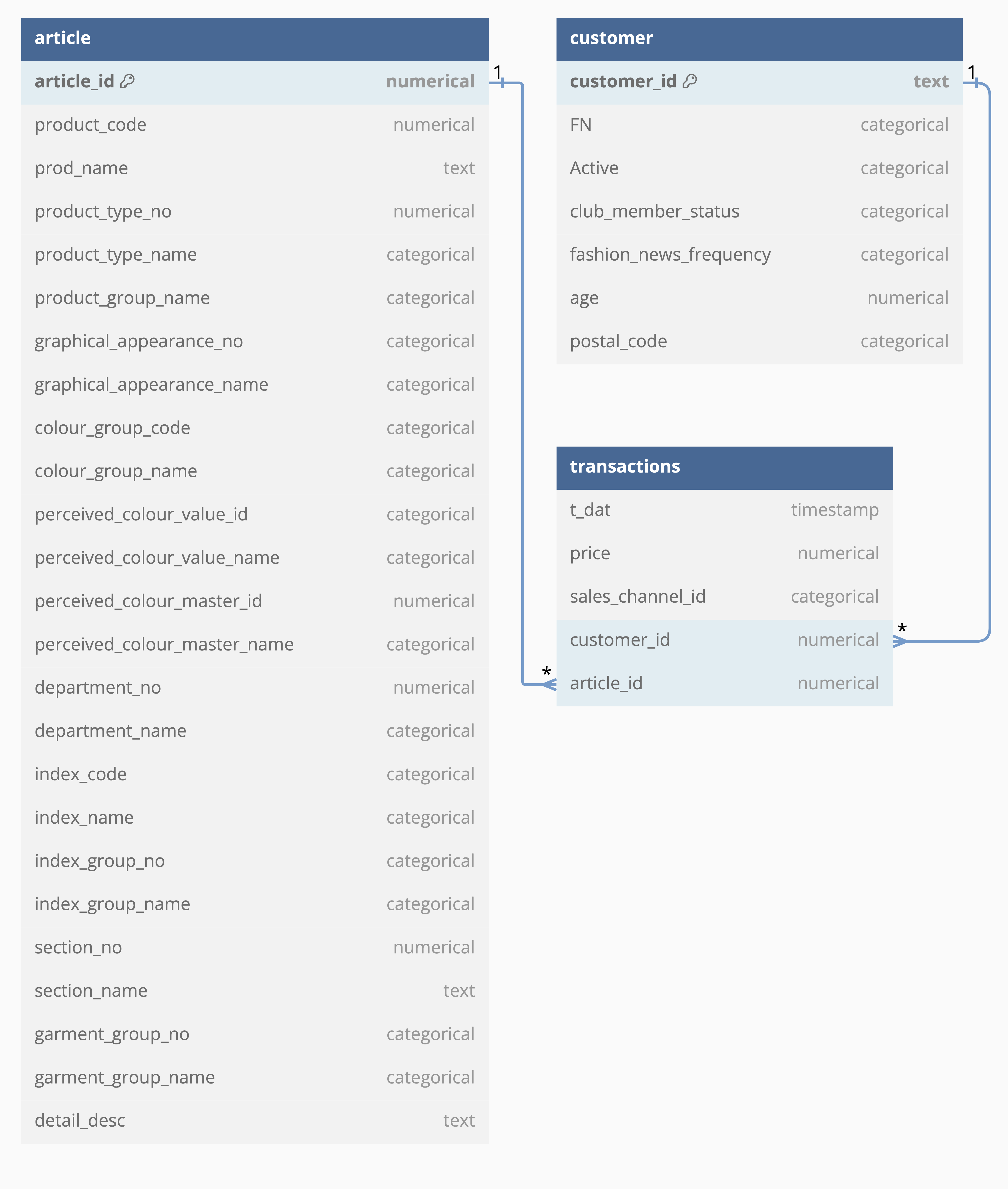}
    \vspace{-0.05in}
    \caption{A database schema for H\&M, which is provided from~\citet{relbench/new}.}
    \label{fig:schema:hm}
\end{figure*}

In this subsection, we provide the additional details for seven tasks (that we design) based on the StackExchange,  Amazon Product Catalog, and H\&M databases. We first report the detailed statistics of the overall datasets in Table~\ref{tab:data_stats}. Also, in Table~\ref{tab:metadata_stats}, we present more fine-grained statistics of each category (column) of the metadata, used for each query. Notably, in this table, we breakdown the metadata features into two categories: `total query' (that includes all the queries in the task) and `non-empty query' (that contains queries with at least one item for each specific metadata category). Lastly, for the schema of each of our considered databases (such as Stack Exchange, Amazon Product Catalog, and H\&M), please refer to Figures~\ref{fig:schema:stack}, \ref{fig:schema:amazon}, and \ref{fig:schema:hm}.

\paragraph{Stack Exchange} 
Recall that, for this database, we design two tasks: \textbf{Answer Retrieval (Any Answer)} and \textbf{Best Answer Retrieval (Best Answer)}. In this paragraph, we describe which specific metadata categories used for query augmentation. At first, for the Answer Retrieval task, we utilize metadata from the post and comment tables. Specifically, we focus on the tags associated with the current question post and the comments on both the current question and the answer posts. For the Best Answer Retrieval task, we utilize metadata from the post, comment, vote, and user tables. The reason why we utilize more categories for this task is because this task is closely related to the personalized retrieval task (for the user who issues the question post); therefore, we focus on constructing the user-specific metadata. Specifically, we use the total comments made by the user, the `aboutme' information of the user, written question and answer posts, and the voted answer posts by the user. Additionally, we include tags from both the current question post and previously asked question posts. For both tasks, we split the queries with their corresponding metadata into training, validation, and test sets, using a corpus of 3,281,834 documents that contain all posts, according to two different settings. In the SplitByUser setting, we randomly sample users in an 8:1:1 ratio from those who have posted questions with answers provided by others. On the other hand, for the SplitByTime setting, we split the datasets based on the creation timestamp of the question posts. Specifically, we create a training set with question posts written before 2019-01-01, a validation set with posts written after 2019-01-01 but before 2020-01-01, and a test set with posts written after 2020-01-01. 

\paragraph{Amazon Product Catalog} 
For this database, we design the \textbf{Future Purchase Retrieval (Future Purchase)} task, where we utilize all the user, product, and review tables. Furthermore, we consider the book reviews written from 2013-01-01 to 2016-01-01 (due to the size of the entire corpus), constructing a document corpus using each product's description. Specifically, we use reviews written in 2013 for the training set, reviews in 2014 for the validation set, and reviews in 2015 for the test set. We then group the reviews written by each customer and randomly sample the customers (since the data before sampling is still very large), selecting 5,000 for the training set, 500 for the validation set, and 500 for the test set. Among two different settings for this task, in the ReviewToProduct setting, each review text (input) is paired with future products (target) that the customer will purchase. For this setting, we incorporate metadata from the previous review text from the review table, and the category, title, and description of both the current and previous products from the product table. In the ProductToProduct setting, we pair the product description of the current review with future products that the customer will buy. We utilize metadata from both the current and previous review texts from the user's review table, along with the category and title of both current and previous products, and the description of the previous products.

\begin{table*}[t!]
\caption{\small Additional results on seven retrieval tasks with Stack Exchange, Amazon Product Catalog, and H\&M databases.}
\vspace{-0.1in}
\label{tab:additional_main}
\small
\centering
\resizebox{\textwidth}{!}{
\renewcommand{\arraystretch}{1.2}
\setlength{\tabcolsep}{4pt}
\begin{tabular}{cl cc cc cc cc cc cc cc cc}
\toprule

& & \multicolumn{4}{c}{\bf StackExchange (Any Answer)} & \multicolumn{4}{c}{\bf StackExchange (Best Answer)} & \multicolumn{4}{c}{\bf Amazon (Future Purchase)} & \multicolumn{2}{c}{\bf H\&M (Future Purchase)} \\
\cmidrule(l{2pt}r{2pt}){3-6} \cmidrule(l{2pt}r{2pt}){7-10} \cmidrule(l{2pt}r{2pt}){11-14} \cmidrule(l{2pt}r{2pt}){15-16}
& & \multicolumn{2}{c}{\bf SplitByUser} & \multicolumn{2}{c}{\bf SplitByTime} & \multicolumn{2}{c}{\bf SplitByUser} & \multicolumn{2}{c}{\bf SplitByTime} & \multicolumn{2}{c}{\bf ReviewToProduct} & \multicolumn{2}{c}{\bf ProductToProduct} & \multicolumn{2}{c}{\bf ProductToProduct}\\
\cmidrule(l{2pt}r{2pt}){3-4} \cmidrule(l{2pt}r{2pt}){5-6} \cmidrule(l{2pt}r{2pt}){7-8}\cmidrule(l{2pt}r{2pt}){9-10} \cmidrule(l{2pt}r{2pt}){11-12} \cmidrule(l{2pt}r{2pt}){13-14} \cmidrule(l{2pt}r{2pt}){15-16}
 & {\bf Method} & MAP & MRR & MAP & MRR & Acc@10 & Acc@50 & Acc@10 & Acc@50 & Acc@1000 & Recall@500 & Acc@1000 & Recall@500 & Acc@100 & Recall@50 \\
\midrule

& BM25-Anserini & 7.10 & 8.61 & 9.99  & 11.01 & 15.50  & 24.58 & 18.55 & 29.14  &  7.77 & 2.78  & 18.39   & 6.53    &  12.63 & 2.49  \\

\midrule \midrule

\multirowcell{7}[-0.0ex][c]{\rotatebox[origin=c]{90}{\textbf{DPR}}} & No Expan. & 23.56 \scriptsize{± 0.03} & 27.86 \scriptsize{± 0.08} & 22.72 \scriptsize{± 0.22} & 25.22 \scriptsize{± 0.24} & 32.75 \scriptsize{± 0.23} & 48.63 \scriptsize{± 0.20} & 35.11 \scriptsize{± 0.60} & 50.96 \scriptsize{± 0.55} & 9.23 \scriptsize{± 0.19} & 1.78 \scriptsize{± 0.27} & 19.73 \scriptsize{± 0.85} & 5.98 \scriptsize{± 0.44}& 14.14 \scriptsize{± 0.88}  & 5.47 \scriptsize{± 0.62} \\

& Expan. w/ LLM & 20.97 \scriptsize{± 0.25} & 24.88 \scriptsize{± 0.30} & 20.12  \scriptsize{± 0.45} & 22.45 \scriptsize{± 0.51} &  28.94 \scriptsize{± 0.85} &  44.05 \scriptsize{± 0.70} &  31.31 \scriptsize{± 0.51} &  46.44 \scriptsize{± 0.30} & 9.35  \scriptsize{± 0.44} & 1.67 \scriptsize{± 0.24} &  19.05  \scriptsize{± 0.22} &  6.05 \scriptsize{± 0.20}& 13.30 \scriptsize{± 0.29}  & 5.12 \scriptsize{± 0.04}  \\

& Expan. w/ LameR & 22.05 \scriptsize{± 0.32} &  26.13 \scriptsize{± 0.35} &  22.11 \scriptsize{± 0.65} &  24.56 \scriptsize{± 0.69} &  30.59 \scriptsize{± 0.45} & 45.15 \scriptsize{± 0.76} & 32.87 \scriptsize{± 0.47} & 48.29 \scriptsize{± 0.50} & 9.05 \scriptsize{± 0.21} & 1.80  \scriptsize{± 0.03 } &  20.57 \scriptsize{± 0.21} &  6.17 \scriptsize{± 0.40} & \textbf{15.66} \scriptsize{± 0.00}  & 5.50 \scriptsize{± 0.01} \\

& Expan. w/ Query & 23.76 \scriptsize{± 0.07} & 28.14 \scriptsize{± 0.09} & 23.67  \scriptsize{± 0.50} & 26.21 \scriptsize{± 0.51} & 32.39  \scriptsize{± 0.47} &  48.74 \scriptsize{± 0.57} &  35.31 \scriptsize{± 0.24} &  51.65 \scriptsize{± 0.37} & 8.57  \scriptsize{± 0.50} & 1.83 \scriptsize{± 0.29} & 21.79   \scriptsize{± 0.21} &  6.59 \scriptsize{± 0.07}& 14.31 \scriptsize{± 0.29}  & 5.50 \scriptsize{± 0.57}  \\

& Expan. w/ User & 23.95 \scriptsize{± 0.20} & 28.14 \scriptsize{± 0.21} &  22.98 \scriptsize{± 0.10} & 25.53 \scriptsize{± 0.12} & 33.57 \scriptsize{± 0.14} &  49.22 \scriptsize{± 0.20} &  35.50 \scriptsize{± 0.35} &  51.68 \scriptsize{± 0.25} &  5.18 \scriptsize{± 0.71} & 1.14 \scriptsize{± 0.11} &  11.25 \scriptsize{± 0.79} & 3.36  \scriptsize{± 0.25}& 14.48 \scriptsize{± 0.58}  & 5.03 \scriptsize{± 0.38}  \\

& Expan. w/ Full & 25.63 \scriptsize{± 0.03} & 30.15 \scriptsize{± 0.07} & 25.16 \scriptsize{± 0.11} & 27.85 \scriptsize{± 0.14} & 31.44 \scriptsize{± 0.47} & 47.13 \scriptsize{± 0.41} & 33.81 \scriptsize{± 0.33} & 49.27 \scriptsize{± 0.27} & 16.10 \scriptsize{± 0.92} & \textbf{4.55} \scriptsize{± 0.24} & 20.74 \scriptsize{± 1.13} & 5.54 \scriptsize{± 0.37}& 7.41 \scriptsize{± 3.79}  & 1.49 \scriptsize{± 0.48} \\

& Expan. w/ Retriever & 25.31 \scriptsize{± 0.04} & 29.79 \scriptsize{± 0.05} & 24.55 \scriptsize{± 0.05} & 27.19 \scriptsize{± 0.09} & 30.98 \scriptsize{± 0.07} & 46.60 \scriptsize{± 0.31} & 33.27 \scriptsize{± 0.15} & 48.72 \scriptsize{± 0.17} & 17.77 \scriptsize{± 0.36} & 4.13 \scriptsize{± 0.21} & 22.65 \scriptsize{± 0.74} & 6.50 \scriptsize{± 0.13}& 12.46 \scriptsize{± 1.46}  & 3.13 \scriptsize{± 0.34} \\

\noalign{\vskip 0.25ex}\cdashline{2-16}\noalign{\vskip 0.75ex}

& DAQu (Ours) & \textbf{27.96} \scriptsize{± 0.23} & \textbf{32.86} \scriptsize{± 0.10} & \textbf{27.58} \scriptsize{± 0.31} & \textbf{30.37} \scriptsize{± 0.35} & \textbf{33.99} \scriptsize{± 0.25} & \textbf{50.05} \scriptsize{± 0.33} & \textbf{36.14} \scriptsize{± 0.42} & \textbf{52.20} \scriptsize{± 0.47} & \textbf{18.01} \scriptsize{± 0.29} & 4.23 \scriptsize{± 0.21} & \textbf{22.68} \scriptsize{± 1.08} & \textbf{7.06} \scriptsize{± 0.15}& 15.49 \scriptsize{± 0.29}  & \textbf{6.62} \scriptsize{± 0.16} \\ \midrule \midrule

\multirowcell{7}[-0.0ex][c]{\rotatebox[origin=c]{90}{\textbf{Contriever}}} & No Expan. & 28.46 \scriptsize{± 0.23} & 33.23 \scriptsize{± 0.19} & 28.38 \scriptsize{± 0.28} & 31.22 \scriptsize{± 0.31} & 39.71 \scriptsize{± 0.42} & 56.13 \scriptsize{± 0.33} & 42.07 \scriptsize{± 0.43} & 57.90 \scriptsize{± 0.20} & 12.62 \scriptsize{± 0.73} & 3.14 \scriptsize{± 0.26} & 21.76 \scriptsize{± 0.37} & 7.65 \scriptsize{± 0.19}& 15.99 \scriptsize{± 0.58}  & 5.69 \scriptsize{± 0.05} \\

& Expan. w/ LLM & 25.75 \scriptsize{± 0.70} & 30.27 \scriptsize{± 0.69} &  25.83 \scriptsize{± 0.16} & 28.49 \scriptsize{± 0.15} & 36.10  \scriptsize{± 0.66} &  51.42 \scriptsize{± 0.29} &  37.42 \scriptsize{± 0.61} &  53.00 \scriptsize{± 0.34} &  12.68 \scriptsize{± 0.18} & 3.25 \scriptsize{± 0.23} &  21.61 \scriptsize{± 0.59} & 7.17  \scriptsize{± 0.36}& 16.16 \scriptsize{± 0.00}  & 5.69 \scriptsize{± 0.28}  \\

& Expan. w/ LameR & 26.14 \scriptsize{± 0.21} &  30.72 \scriptsize{± 0.14} &  25.96 \scriptsize{± 0.04} &  28.74 \scriptsize{± 0.00} & 37.07 \scriptsize{± 0.22} &  52.81 \scriptsize{± 0.02} &  37.50 \scriptsize{± 0.35} & 52.15  \scriptsize{± 0.70} &   10.09 \scriptsize{± 0.15} &  3.07 \scriptsize{± 0.31} &  21.22 \scriptsize{± 0.21} &  6.94 \scriptsize{± 0.05 } & 15.32 \scriptsize{± 0.29}  & 5.75 \scriptsize{± 0.01} \\

& Expan. w/ Query & 28.15 \scriptsize{± 0.34} & 32.99 \scriptsize{± 0.41} &  28.58 \scriptsize{± 0.13} & 31.43 \scriptsize{± 0.11} &  37.43 \scriptsize{± 0.26} &  54.99 \scriptsize{± 0.47} & 41.11  \scriptsize{± 0.24} &  57.72 \scriptsize{± 0.14} &  13.39 \scriptsize{± 0.92} & 3.29 \scriptsize{± 0.12} &  22.86  \scriptsize{± 0.29} &  7.74 \scriptsize{± 0.28}& 15.91 \scriptsize{± 0.36}  & 5.80 \scriptsize{± 0.12}  \\

& Expan. w/ User & 28.88 \scriptsize{± 0.21} & 33.63 \scriptsize{± 0.21} & 28.07 \scriptsize{± 0.32} & 30.94 \scriptsize{± 0.29} &  39.32 \scriptsize{± 0.17} &  55.92 \scriptsize{± 0.28} &  42.30 \scriptsize{± 0.42} & 57.64 \scriptsize{± 0.56} &  8.57 \scriptsize{± 0.52} & 1.57 \scriptsize{± 0.23} &  11.43  \scriptsize{± 0.67} &  3.16 \scriptsize{± 0.31}& 12.29 \scriptsize{± 0.29}  & 4.14 \scriptsize{± 0.46}  \\

& Expan. w/ Full & 31.06 \scriptsize{± 0.16} & 36.12 \scriptsize{± 0.12} & 30.12 \scriptsize{± 0.08} & 33.14 \scriptsize{± 0.08} & 39.28 \scriptsize{± 0.35} & 56.04 \scriptsize{± 0.43} & 41.32 \scriptsize{± 0.15} & 57.33 \scriptsize{± 0.53} & 22.65 \scriptsize{± 0.67} & 7.07 \scriptsize{± 0.14} & 23.60 \scriptsize{± 0.88} & 7.14 \scriptsize{± 0.36}& 6.90 \scriptsize{± 0.58}  & 1.34 \scriptsize{± 0.01} \\

& Expan. w/ Retriever & 30.82 \scriptsize{± 0.19} & 35.76 \scriptsize{± 0.22} & 30.30 \scriptsize{± 0.32} & 33.24 \scriptsize{± 0.35} & 38.09 \scriptsize{± 0.50} & 54.56 \scriptsize{± 0.25} & 40.79 \scriptsize{± 0.45} & 56.42 \scriptsize{± 0.41} & 22.62 \scriptsize{± 0.22} & 5.42 \scriptsize{± 0.44} & 22.62 \scriptsize{± 0.22} & 7.44 \scriptsize{± 0.04}& 14.98 \scriptsize{± 0.58}  & 4.29 \scriptsize{± 0.13} \\

\noalign{\vskip 0.25ex}\cdashline{2-16}\noalign{\vskip 0.75ex}

& DAQu (Ours) & \textbf{35.00} \scriptsize{± 0.33} & \textbf{40.55} \scriptsize{± 0.41} & \textbf{34.96} \scriptsize{± 0.53} & \textbf{38.07} \scriptsize{± 0.57} & \textbf{40.50} \scriptsize{± 0.16} & \textbf{57.59} \scriptsize{± 0.58} & \textbf{42.53} \scriptsize{± 0.06} & \textbf{58.48} \scriptsize{± 0.51} & \textbf{25.65} \scriptsize{± 0.44} & \textbf{7.10} \scriptsize{± 0.29} & \textbf{25.36} \scriptsize{± 0.50} & \textbf{8.31} \scriptsize{± 0.23}& \textbf{17.17} \scriptsize{± 1.43}  & \textbf{5.81} \scriptsize{± 0.49} \\ \midrule \midrule

\multirowcell{7}[-0.0ex][c]{\rotatebox[origin=c]{90}{\textbf{BGE-M3}}} 

& No Expan. &  26.23 \scriptsize{± 0.49} & 30.73 \scriptsize{± 0.62} & 25.72 \scriptsize{± 0.30} & 28.32 \scriptsize{± 0.29} &  35.14 \scriptsize{± 0.78} & 51.30 \scriptsize{± 0.12} & 35.44 \scriptsize{± 0.22} & 50.36 \scriptsize{± 0.53} & 11.52 \scriptsize{± 0.15} & 2.62 \scriptsize{± 0.06} & 21.34 \scriptsize{± 0.15} & 6.61 \scriptsize{± 0.01}& 14.98  \scriptsize{± 0.58}  & 5.46  \scriptsize{± 0.03}  \\

& Expan. w/ LLM & 25.14 \scriptsize{± 0.21} & 29.65 \scriptsize{± 0.19} & 25.20 \scriptsize{± 0.13} & 27.89 \scriptsize{± 0.09} &  30.03 \scriptsize{± 0.30} & 44.76 \scriptsize{± 0.78} & 31.18 \scriptsize{± 0.20} & 45.08 \scriptsize{± 0.36} & 11.67 \scriptsize{± 1.29} & 2.50 \scriptsize{± 0.47} & 20.60 \scriptsize{± 0.36} & 6.35 \scriptsize{± 0.06}& 15.15 \scriptsize{± 0.00}  & 5.52 \scriptsize{± 0.17}  \\

& Expan. w/ LameR & 25.83 \scriptsize{± 0.37} &  30.29 \scriptsize{± 0.42} &  25.72 \scriptsize{± 0.20} & 28.38  \scriptsize{± 0.28} & 31.31 \scriptsize{± 0.87} & 45.84  \scriptsize{± 0.37} & 32.28 \scriptsize{± 0.47 } &  46.41 \scriptsize{± 0.37 } &  10.51 \scriptsize{± 1.13} &  2.44 \scriptsize{± 0.47} &  19.49 \scriptsize{± 0.10 } & 6.54  \scriptsize{± 0.23} & 15.66 \scriptsize{± 0.87}  & 5.16 \scriptsize{± 0.03} \\

& Expan. w/ Query & 25.86 \scriptsize{± 0.57} & 30.25 \scriptsize{± 0.73} & 26.48 \scriptsize{± 0.41} & 29.15 \scriptsize{± 0.43} & 36.39  \scriptsize{± 0.31} & 52.76 \scriptsize{± 0.89} & 35.90 \scriptsize{± 0.74} & 51.93 \scriptsize{± 0.73} & 11.16 \scriptsize{± 0.46} & 2.41 \scriptsize{± 0.18} & 20.60 \scriptsize{± 0.05} & 6.55 \scriptsize{± 0.15}& 16.33 \scriptsize{± 0.29}  & 5.62 \scriptsize{± 0.15}  \\

& Expan. w/ User & 27.41 \scriptsize{± 0.36} & 31.98 \scriptsize{± 0.38} & 27.66 \scriptsize{± 0.11} & 30.41 \scriptsize{± 0.11} & 36.29  \scriptsize{± 0.96} & 52.02 \scriptsize{± 1.19} & 35.91 \scriptsize{± 0.55} & 51.38 \scriptsize{± 0.54} & 6.34 \scriptsize{± 1.86} & 1.33 \scriptsize{± 0.19} & 15.33 \scriptsize{± 0.10} & 3.77 \scriptsize{± 0.30}& 15.99 \scriptsize{± 0.58}  & 5.62 \scriptsize{± 0.03}  \\

& Expan. w/ Full & 27.35 \scriptsize{± 0.17} & 32.03 \scriptsize{± 0.16} & 27.06 \scriptsize{± 0.83} & 29.78 \scriptsize{± 0.92} & 35.94 \scriptsize{± 0.27} & 51.27 \scriptsize{± 1.04} & 35.46 \scriptsize{± 0.05} & 50.31 \scriptsize{± 0.30} & 17.89 \scriptsize{± 0.82} & \textbf{5.39} \scriptsize{± 0.31} & 20.98 \scriptsize{± 2.78} & 5.76 \scriptsize{± 0.61}& 6.40 \scriptsize{± 0.58}  & 1.36 \scriptsize{± 0.04}  \\

& Expan. w/ Retriever & 27.91 \scriptsize{± 0.49} & 32.59 \scriptsize{± 0.44} & 27.43 \scriptsize{± 0.16} & 30.14 \scriptsize{± 0.20} & 35.84  \scriptsize{± 0.02} & 51.02 \scriptsize{± 0.32} & 34.22 \scriptsize{± 0.55} & 49.31 \scriptsize{± 0.89} & 17.53 \scriptsize{± 0.05} & 4.29 \scriptsize{± 0.11} & 23.27 \scriptsize{± 0.36} & 6.34 \scriptsize{± 0.40}& 15.99 \scriptsize{± 0.58}  & 4.50 \scriptsize{± 0.28}  \\

\noalign{\vskip 0.25ex}\cdashline{2-16}\noalign{\vskip 0.75ex}

& DAQu (Ours) & \textbf{30.26} \scriptsize{± 0.30} & \textbf{35.05} \scriptsize{± 0.30} & \textbf{30.17} \scriptsize{± 0.38} & \textbf{33.00} \scriptsize{± 0.43} &  \textbf{38.26} \scriptsize{± 1.03} & \textbf{54.09} \scriptsize{± 0.54} & \textbf{36.56} \scriptsize{± 0.22} &  \textbf{52.05} \scriptsize{± 0.01} & \textbf{20.30} \scriptsize{± 1.34} & 4.78 \scriptsize{± 0.51} & \textbf{23.36} \scriptsize{± 0.21} & \textbf{6.86} \scriptsize{± 0.15}& \textbf{17.51} \scriptsize{± 0.29}  & \textbf{5.81} \scriptsize{± 0.03}  \\

\bottomrule

\end{tabular}
}
\vspace{-0.1in}
\end{table*}
\begin{table}[t!]
\caption{\small Metadata statistics (Best Answer, SplitByUser).}
\vspace{-0.1in}
\label{tab:long_metadata}
\small
\centering
\resizebox{0.475\textwidth}{!}{
\renewcommand{\arraystretch}{0.975}
\begin{tabular}{lcc}
\toprule
\textbf{Metadata Category} & \textbf{Train (Avg Words per Query)} & \textbf{Test (Avg Words per Query)} \\

\midrule
\midrule

\multirowcell{1}[-0.0ex][l]{\textbf{Question Posts}}
& 2,459.08 & 1,849.05  \\

\multirowcell{1}[-0.0ex][l]{\textbf{Answer Posts}}
& 3,690.50 & 2,934.33 \\

\multirowcell{1}[-0.0ex][l]{\textbf{Accepted Answers}} 
& 1,717.59 & 1,493.52  \\

\multirowcell{1}[-0.0ex][l]{\textbf{Comments}} 
& 2,844.51 & 3,169.55  \\

\multirowcell{1}[-0.0ex][l]{\textbf{About Me}} 
& 9.04 & 10.33  \\

\multirowcell{1}[-0.0ex][l]{\textbf{Current Tags}} 
& 3.06 & 3.08  \\

\multirowcell{1}[-0.0ex][l]{\textbf{Previous Tags}} 
& 48.36	 & 41.59 \\

\noalign{\vskip 0.25ex}\cdashline{1-3}\noalign{\vskip 0.75ex}

\multirowcell{1}[-0.0ex][l]{\textbf{Total Words}} 
& \textbf{10,772.14} &  \textbf{9,501.45} \\

\multirowcell{1}[-0.0ex][l]{\textbf{Longest Metadata}} 
& \textbf{307,016} & \textbf{439,969}  \\

\bottomrule
\end{tabular}
}
\end{table}

\paragraph{H\&M}
Similar to the Amazon Product Catalog setting, our goal is to predict the future products a customer will purchase by leveraging metadata from both current and previous products, utilizing information from all the user, article, and transaction tables. To achieve this, we consider purchases made between 2020-01-01 and 2020-09-14, using data from 2020-01-01 to 2020-07-01 as the training set, 2020-07-01 to 2020-08-01 as the validation set, and 2020-08-01 to 2020-09-14 as the test set.

\subsection{Models}
For DPR~\cite{dpr}, we follow the implementation by~\citet{beir}. For Contriever~\cite{contriever}, we further train it from its available checkpoint while using the same architecture as DPR. For a fair comparison, we fix the number of epochs across the same retrieval models for each task and report the average of the three different runs for every model. We use A100 GPU clusters for conducting experiments.

\section{Experimental Results}

\subsection{Results with Other Metrics}\label{sup:additional_result}
In addition to our main results in Table~\ref{tab:main}, we provide the results with other retrieval metrics in Table~\ref{tab:additional_main}. From this, similar to the results in Table~\ref{tab:main}, we also observe that our DAQu shows remarkable performance improvements in diverse scenarios. 

\subsection{Metadata Length Challenges}
Our graph-based set-encoding strategy is particularly beneficial when dealing with concatenated textual metadata that may be very long for the encoder to handle. As shown in the metadata statistics in Table~\ref{tab:long_metadata}, the concatenated metadata often results in substantial word counts, with some cases exceeding the token limits of commonly used LLMs, making them impractical for direct processing. Moreover, even when token limits are not exceeded, processing such long contexts can lead to significant computational overhead. These challenges further emphasize the advantages of our graph-based set-encoding approach, which efficiently encodes metadata while preserving its structure and hierarchy.

\begin{table}[t!]
\caption{\small Results for Expan. w/ Full with a special token for each metadata category (DPR, Any Answer, SplitByTime).}
\vspace{-0.05in}
\label{tab:naive_specialtoken}
\small
\centering
\resizebox{0.475\textwidth}{!}{
\renewcommand{\arraystretch}{0.975}
\begin{tabular}{lcc}
\toprule
\textbf{Method} & \textbf{Recall@10} & \textbf{Acc@100} \\

\midrule
\midrule

\multirowcell{1}[-0.0ex][l]{\textbf{No Expan.}}
& 35.46	 & 64.48 \\

\multirowcell{1}[-0.0ex][l]{\textbf{Expan. w/ Full}} 
& 38.75 & 67.37 \\

\multirowcell{1}[-0.0ex][l]{\textbf{Expan. w/ Full (w/ Special Tokens)}} 
& 38.31  & 67.35 \\

\noalign{\vskip 0.25ex}\cdashline{1-3}\noalign{\vskip 0.75ex}

\multirowcell{1}[-0.0ex][l]{\textbf{DAQu (Ours)}} 
& \textbf{41.67} & \textbf{71.72} \\

\bottomrule
\end{tabular}
}
\end{table}

\subsection{Metadata Expansion with Special Token}
To evaluate the impact of using special tokens for differentiating metadata categories on retrieval performance for the Full Metadata Expansion baseline (which concatenates a given query with all metadata terms), we extend it by including special tokens for metadata differentiation. As shown in Table~\ref{tab:naive_specialtoken}, the inclusion of special tokens has minimal effect on performance, with Full Metadata Expansion achieving comparable retrieval results regardless of their use.

\begin{table}[t!]
\caption{\small Results with Recall@20 and Acc@20, for Table~\ref{tab:ablation}.}
\vspace{-0.05in}
\small 
\centering
\resizebox{0.495\textwidth}{!}{
\renewcommand{\arraystretch}{0.95}
\begin{tabular}{llcc}
\toprule
& \textbf{Method} & \textbf{Recall@20} & \textbf{Acc@20} \\

\midrule

& \multirowcell{1}[-0.0ex][l]{\textbf{BM25-Anserini}}
& 14.43  & 17.43   \\

\midrule
\midrule

\multirow{7}{*}{\rotatebox[origin=c]{90}{\scriptsize{\textbf{DPR}}}}  

& \textbf{No Expan.} &  43.09 \scriptsize{± 0.21} & 50.35 \scriptsize{± 0.29}    \\

& {\textbf{Expan. w/ LLM}}
& 39.12 \scriptsize{± 0.33} & 45.97 \scriptsize{± 0.33} \\

& {\textbf{Expan. w/ Query}}
& 44.04 \scriptsize{± 0.33} & 51.28 \scriptsize{± 0.30} \\

& {\textbf{Expan. w/ User}}
& 43.31 \scriptsize{± 0.07} & 50.49 \scriptsize{± 0.13 }\\

& {\textbf{Expan. w/ Full}}
&  46.20 \scriptsize{± 0.07 }&  53.66 \scriptsize{± 0.09 } \\

& {\textbf{Expan. w/ Retriever}}
& 45.70 \scriptsize{± 0.03}	  &   53.05 \scriptsize{± 0.05} \\

\cdashline{2-4}\noalign{\vskip 0.75ex}

& {\textbf{DAQu (Ours)}}
& \textbf{49.54 }\scriptsize{± 0.23} &   \textbf{57.13} \scriptsize{± 0.12} \\

\midrule
\midrule

\multirow{7}{*}{\rotatebox[origin=c]{90}{\scriptsize{\textbf{Contriever}}}}  & \textbf{No Expan.}  & 49.20 \scriptsize{± 0.26} & 56.79 \scriptsize{± 0.28 } \\

& {\textbf{Expan. w/ LLM}}
& 45.24 \scriptsize{± 0.67} & 52.64 \scriptsize{± 0.71} \\

& {\textbf{Expan. w/ Query}}
& 49.73 \scriptsize{± 0.38 }& 57.49 \scriptsize{± 0.48} \\

& {\textbf{Expan. w/ User}}
& 50.00 \scriptsize{± 0.31} & 57.45 \scriptsize{± 0.46} \\

& {\textbf{Expan. w/ Full}}
& 52.57 \scriptsize{± 0.12}	  &  60.26 \scriptsize{± 0.10}  \\

& {\textbf{Expan. w/ Retriever}}
&  52.23 \scriptsize{± 0.24}	 &  59.78 \scriptsize{± 0.25}  \\

\cdashline{2-4}\noalign{\vskip 0.75ex}

& {\textbf{DAQu (Ours)}}
&  \textbf{57.33} \scriptsize{± 0.07}	 &  \textbf{65.05} \scriptsize{± 0.09}  \\

\midrule
\midrule

\multirow{7}{*}{\rotatebox[origin=c]{90}{\scriptsize{\textbf{BGE-M3}}}}  

& \textbf{No Expan.}
& 47.02 \scriptsize{± 0.44} & 54.38   \scriptsize{± 0.47} \\

& {\textbf{Expan. w/ LLM}}
& 44.08  \scriptsize{± 0.20} & 51.43   \scriptsize{± 0.24} \\

& {\textbf{Expan. w/ Query}}
& 47.34  \scriptsize{± 1.03} &  54.83  \scriptsize{± 1.19} \\

& {\textbf{Expan. w/ User}}
& 48.68  \scriptsize{± 0.15} &  56.08  \scriptsize{± 0.12} \\

& {\textbf{Expan. w/ Full}}
& 48.83  \scriptsize{± 0.02} & 56.24   \scriptsize{± 0.02} \\

& {\textbf{Expan. w/ Retriever}}
& 49.07  \scriptsize{± 0.49} & 56.47   \scriptsize{± 0.67} \\

\cdashline{2-4}\noalign{\vskip 0.75ex}

& {\textbf{DAQu (Ours)}}
& \textbf{52.33}  \scriptsize{± 0.04} &  \textbf{60.00}  \scriptsize{± 0.19} \\

\bottomrule
\end{tabular}
}
\vspace{-0.05in}
\label{tab:same_metric}
\end{table}

\subsection{Results with Consistent Metrics}
In addition to reporting results with diverse metrics to demonstrate the effectiveness of the proposed method across various evaluation criteria, we also provide the results in Table~\ref{tab:same_metric} using the same metrics as in Table~\ref{tab:ablation}. As shown in Table~\ref{tab:same_metric}, these results are consistent with our previous findings, further confirming that our DAQu framework significantly outperforms the baseline methods. 

\subsection{Results of LameR with BM25}
Since LameR is originally designed for sparse retrieval settings (yet we adopt it with the dense retriever to compare against our DAQu framework tailored for dense retrieval), we further explore the variant of LameR with BM25 in Table~\ref{tab:lamer}. From this, we find that although LameR provides some improvements coupled with the BM25 retriever, it still lags significantly behind DAQu, which leverages structured metadata in the latent space. These results indicate the importance of dense retrieval, especially in tasks where understanding nuanced relationships in metadata is crucial.

\begin{table}[t!]
\caption{\small Comparison between BM25-based LameR and our DAQu (with DPR) on the Any Answer and Best Answer tasks.}
\vspace{-0.05in}
\label{tab:lamer}
\small
\centering
\resizebox{0.475\textwidth}{!}{
\renewcommand{\arraystretch}{1.0}
\begin{tabular}{lccc}
\toprule
\textbf{Method} & \textbf{MRR} & \textbf{Acc@20}  & \textbf{Acc@100} \\

\midrule
\midrule

\multicolumn{4}{c}{Any Answer, SplitByUser} \\ \midrule

\multirowcell{1}[-0.0ex][l]{\textbf{BM25}}
& 8.61 & 17.43 & 28.33 \\

\multirowcell{1}[-0.0ex][l]{\textbf{BM25 w/ LameR}} 
&  10.66 & 21.30  & 35.14 \\

\noalign{\vskip 0.25ex}\cdashline{1-4}\noalign{\vskip 0.75ex}

\multirowcell{1}[-0.0ex][l]{\textbf{DAQu (Ours)}} 
& \textbf{32.86} & \textbf{57.13} & \textbf{74.11} \\

\midrule
\midrule

\multicolumn{4}{c}{Best Answer, SplitByUser} \\ \midrule

\multirowcell{1}[-0.0ex][l]{\textbf{BM25}}
& 9.64 & 19.42 & 29.49 \\

\multirowcell{1}[-0.0ex][l]{\textbf{BM25 w/ LameR}} 
& 11.70  & 23.53  & 36.48 \\

\noalign{\vskip 0.25ex}\cdashline{1-4}\noalign{\vskip 0.75ex}

\multirowcell{1}[-0.0ex][l]{\textbf{DAQu (Ours)}} 
& \textbf{22.05} & \textbf{40.40 } & \textbf{ 57.81} \\

\bottomrule
\end{tabular}
}
\end{table}

\begin{table}[t!]
\caption{\small Results of using Contriever in the `Expan. w/ Retriever' baseline (Any Answer, SplitByUser).}
\vspace{-0.05in}
\label{tab:expan_w_contriever}
\small
\centering
\resizebox{0.475\textwidth}{!}{
\renewcommand{\arraystretch}{0.975}
\begin{tabular}{lcc}
\toprule
\textbf{Method}  & \textbf{Recall@10}  & \textbf{Acc@100} \\

\midrule
\midrule

\multirowcell{1}[-0.0ex][l]{\textbf{No Expan.}}
& 42.08  & 73.21  \\

\multirowcell{1}[-0.0ex][l]{\textbf{Expan. w/ Retriever (BM25)}}
& 44.69  &  75.52 \\

\multirowcell{1}[-0.0ex][l]{\textbf{Expan. w/ Retriever (Contriever)}} 
& 44.66 & 76.12 \\

\noalign{\vskip 0.25ex}\cdashline{1-3}\noalign{\vskip 0.75ex}

\multirowcell{1}[-0.0ex][l]{\textbf{DAQu (Ours)}} 
& \textbf{49.74} & \textbf{80.27} \\

\bottomrule
\end{tabular}
}
\end{table}

\subsection{Results of Expan. w/ Retriever Variant}
For the `Expan. w/ Retriever' baseline, following~\citet{Deng2021TowardPA}, we adopt a BM25 model to select metadata terms most relevant to the query and append only those selected terms. To further examine the impact of the retriever used for metadata selection, we replace BM25 with a dense retriever, Contriever. As shown in Table~\ref{tab:expan_w_contriever}, while both retriever-based expansion methods offer moderate gains over the no-expansion baseline, DAQu consistently and significantly outperforms both of them. This highlights the effectiveness of integrating metadata in the latent space, rather than relying solely on term-level expansion.

\subsection{Impact of Query-Associated Metadata on Retrieval Diversity}
\label{appendix:diversity}
While the primary focus of our work is to explore how structured metadata from relational databases can enhance retrieval relevance in scenarios where such metadata is available, we further examine whether the query-associated metadata used for query augmentation leads to repetitive or overly narrow retrieval. Specifically, we measure the average overlap ratio and the Jaccard similarity between retrieved results for different queries from the same user, averaged across all users. We then observe the negligible redundancy: the Jaccard similarity is 0.0096 for the no-expansion baseline and 0.0104 for our method, and the overlap ratio is 0.0084 and 0.0091, respectively, indicating that the likelihood that a user receives an overlapping element (retrieved) across queries is below 1\% and suggesting that the retrieval results remain diverse even after metadata augmentation. Nevertheless, while negligible in our experimental results, the metadata can introduce subtle bias depending on the context, and we leave exploring it as future work.

\subsection{Case Study}
\label{appendix:case}
We conduct a case study to qualitatively compare the effectiveness of our DAQu against the baseline query augmentation methods, provided in Table~\ref{tab:case_study}. The first example from the Any Answer retrieval task with the SplitByTime setting presents retrieval results for a user query: selecting optimal activation and loss functions when training an autoencoder on the MNIST dataset. Notably, the challenge here is several important keywords with query-relevant information, such as BCE and MSE, are missing from the original user query. While the baseline expansion models can include such keywords, which can lead to a higher rank of the relevant document (Full Metadata Expansion), Expansion with Retriever results in a lower rank than even No Expansion, due to the exclusion of another essential term, `Keras'. In contrast, DAQu achieves the highest rank among all baselines, indicating that our method effectively augments all essential information with the metadata representation, by utilizing diverse helpful information sources in a relational database. Similarly, for the Best Answer retrieval task with the SplitByTime setting, given a query such as when normalization or standardization is appropriate, the best answer post explains such cases in terms of `transformation methods.' Here, our DAQu, which can incorporate the relevant term `log transformation' from the metadata into the query representation, achieves the highest rank. Finally, for the Future Product retrieval task, a user purchased the book `Kindergarten-Grade 3' for their children. In addition, this user's metadata includes information on several previous purchases tagged `Children's Books.' In this example, while the No Expansion baseline effectively retrieves the future product with a higher rank, Full Metadata Expansion and Expansion with Retriever do not perform well, suggesting that augmenting metadata with text level adds noise to the retrieval process. Meanwhile, our DAQu effectively exploits only the valuable information on the latent space, achieving the highest rank among all models.

\subsection{Error Analysis}
\label{appendix:error}
As the datasets used in our experiments are collected from real-world applications (e.g., Amazon or StackExchange), the associated metadata is naturally noisy, incomplete, or sometimes weakly relevant to the current query. Then, to better illustrate how our method behaves under such realistic and noisy conditions, we include an error case study in Table~\ref{tab:error_case_study}. As shown in this example, although the query is about evaluating probabilistic predictions, the metadata includes loosely related or distractive content, such as a vague recommendation link and general comments. As a result, the retriever with full metadata expansion (Expan. w/ Full) ranks the correct document much lower (Rank 24). However, our DAQu framework still ranks the correct answer at Rank 4, demonstrating robustness to noise and the ability to effectively utilize metadata signals. This example highlights that while noisy metadata can introduce challenges, DAQu remains effective by learning to selectively incorporate relevant signals (at the representation level), rather than naively aggregating all available metadata in text.

\begin{table*}[t!]
\caption{Distribution of the metadata features per query for each metadata category for various retrieval tasks.}
\vspace{-0.1in}
\label{tab:metadata_stats}
\small
\centering
\resizebox{.8\textwidth}{!}{
\renewcommand{\arraystretch}{1.0}
\begin{tabular}{l l ccc c ccc}
\toprule
& & \multicolumn{3}{c}{\textbf{Total Query}} & & \multicolumn{3}{c}{\textbf{Non Empty Query}} \\ \cmidrule(l{2pt}r{2pt}){3-5} \cmidrule(l{2pt}r{2pt}){7-9}  
\textbf{Setting} & \textbf{Metadata Category} & \textbf{Training} & \textbf{Valid} & \textbf{Test} & &\textbf{Training}& \textbf{Valid}& \textbf{Test} \\ \midrule
\multicolumn{9}{c}{\textit{StackExchange - Any Answer}} \\ \midrule
\multirowcell{3}[-0.0ex][l]{SplitByUser} & \textsf{comments\_in\_question} & 1.96 & 1.95 & 1.94 & & 3.35 & 3.37 & 3.31\\ 
 & \textsf{comments\_in\_answers} & 2.31 & 2.45 & 2.31 & & 3.96 & 4.14 & 3.99 \\
& \textsf{tags} & 3.00 & 3.04 & 3.01 & & 3.00 & 3.04 & 3.01 \\  \noalign{\vskip 0.25ex}\cdashline{1-9}\noalign{\vskip 0.75ex}
\multirowcell{3}[-0.0ex][l]{SplitByTime} & \textsf{comments\_in\_question} & 2.03 & 1.69 & 1.63 & & 3.38 & 3.19 & 3.26\\ 
& \textsf{comments\_in\_answers} & 2.43 & 1.89 & 2.08 & & 4.09 & 3.46 & 3.71\\
& \textsf{tags} & 2.97 & 3.06 & 3.23 & & 2.97 & 3.06 & 3.23\\ \midrule
\multicolumn{9}{c}{\textit{StackExchange - Best Answer}} \\ \midrule
\multirowcell{7}[-0.0ex][l]{SplitByUser} & \textsf{question\_posts} & 14.52 & 22.15 & 12.42 && 18.18 & 27.07 & 15.77 \\ 
 & \textsf{answer\_posts} & 19.77 & 24.25 & 13.47 & & 44.79 & 55.18 & 30.74\\
 & \textsf{accepted\_answers} & 7.41 & 13.41 & 6.25 & & 10.91 & 18.68 & 9.41\\
& \textsf{comments} & 81.28 & 122.02 & 84.92 & & 92.86 & 137.92 & 97.46\\  
& \textsf{aboutme} & 0.33 & 0.31 & 0.33 & & 1.00 & 1.00 & 1.00\\  
& \textsf{current\_tags} & 3.06 & 2.99 & 3.08 & & 3.06 & 2.99 & 3.08 \\  
& \textsf{previous\_tags} & 48.36 & 66.99 & 41.59 & & 48.36 & 66.99 & 41.59 \\  
\noalign{\vskip 0.25ex}\cdashline{1-9}\noalign{\vskip 0.75ex}
\multirowcell{7}[-0.0ex][l]{SplitByTime} & \textsf{question\_posts} & 6.52 & 7.04 & 9.96 & & 10.46 & 11.25 & 14.94\\ 
& \textsf{answer\_posts} & 7.82 & 9.35 & 11.15 && 27.47 & 38.98 & 42.83\\
& \textsf{accepted\_answers} & 3.82 & 3.67 & 5.36 && 7.29 & 7.21 & 9.77\\  
& \textsf{comments} & 31.09 & 38.59 & 49.44 && 54.32 & 67.36 & 81.55\\  
& \textsf{aboutme} & 0.34 & 0.29 & 0.28 && 1 & 1 & 1\\  
& \textsf{current\_tags} & 3.02 & 3.10 & 3.25 && 3.02 & 3.10 & 3.25\\  
& \textsf{previous\_tags} & 19.52 & 21.71 & 32.33 && 31.31 & 34.70 & 48.52\\  \midrule
\multicolumn{9}{c}{\textit{Amazon Product Catalog}} \\ \midrule
\multirowcell{7}[-0.0ex][l]{ReviewToProduct} & \textsf{previous\_review\_text} & 8.22 & 6.97 & 15.05  & & 11.22 & 8.94 & 17.52\\ 
 & \textsf{current\_product\_category} & 2.90 & 2.91 & 2.86 && 2.99 & 3.00 & 2.99\\
 & \textsf{current\_product\_title} & 1.00 & 1.00 & 1.00 && 1.00 & 1.00 & 1.00\\
& \textsf{current\_product\_description} & 1.00 & 1.00 & 1.00 && 1.00 & 1.00 & 1.00\\  
& \textsf{previous\_product\_category} & 23.96 & 20.34 & 44.16 && 33.01 & 26.39 & 52.68\\  
& \textsf{previous\_product\_category} & 8.22 & 6.97 & 15.05 && 11.22 & 8.94 & 17.52\\  
& \textsf{previous\_product\_description} & 8.22 & 6.97 & 15.05 && 11.22 & 8.94 & 17.52\\  
\noalign{\vskip 0.25ex}\cdashline{1-9}\noalign{\vskip 0.75ex}
\multirowcell{7}[-0.0ex][l]{ProductToProduct} & \textsf{previous\_review\_text} & 8.22 & 6.97 & 15.05  & & 11.22 & 8.94 & 17.52\\ 
 & \textsf{current\_product\_category} & 2.90 & 2.91 & 2.86 && 2.99 & 3.00 & 2.99\\
 & \textsf{current\_product\_title} & 1.00 & 1.00 & 1.00 && 1.00 & 1.00 & 1.00\\
& \textsf{current\_product\_description} & 1.00 & 1.00 & 1.00 && 1.00 & 1.00 & 1.00\\  
& \textsf{previous\_product\_category} & 23.96 & 20.34 & 44.16 && 33.01 & 26.39 & 52.68\\  
& \textsf{previous\_product\_category} & 8.22 & 6.97 & 15.05 && 11.22 & 8.94 & 17.52\\  
& \textsf{previous\_product\_description} & 8.22 & 6.97 & 15.05 && 11.22 & 8.94 & 17.52\\ \midrule
\multicolumn{9}{c}{\textit{H\&M}} \\ \midrule
\multirowcell{11}[-0.0ex][l]{ProductToProduct} 
& \textsf{customer\_age} & 1.00 & 1.00 & 1.00 && 1.00 & 1.00 & 1.00 \\
& \textsf{customer\_fashion\_news\_frequency} & 1.00 & 1.00 & 1.00 && 1.00 & 1.00 & 1.00 \\
& \textsf{previous\_product\_description} & 18.56 & 45.04 & 36.44 && 20.57 & 46.77 & 37.28 \\
& \textsf{current\_product\_product\_type\_name} & 1.00 & 1.00 & 1.00 && 1.00 & 1.00 & 1.00 \\
& \textsf{current\_product\_product\_group\_name} & 1.00 & 1.00 & 1.00 && 1.00 & 1.00 & 1.00 \\
& \textsf{current\_product\_perceived\_colour\_master\_name} & 1.00 & 1.00 & 1.00 && 1.00 & 1.00 & 1.00 \\
& \textsf{current\_product\_department\_name} & 1.00 & 1.00 & 1.00 && 1.00 & 1.00 & 1.00 \\
& \textsf{current\_product\_index\_name} & 1.00 & 1.00 & 1.00 && 1.00 & 1.00 & 1.00 \\
& \textsf{current\_product\_index\_group\_name} & 1.00 & 1.00 & 1.00 && 1.00 & 1.00 & 1.00 \\
& \textsf{current\_product\_section\_name} & 1.00 & 1.00 & 1.00 && 1.00 & 1.00 & 1.00 \\
& \textsf{current\_product\_garment\_group\_name} & 1.00 & 1.00 & 1.00 && 1.00 & 1.00 & 1.00 \\

\bottomrule

\end{tabular}
}
\end{table*}

\begin{table*}[t!]
\caption{Case study on three retrieval tasks. In response to the query from the user, notable terms in the Metadata and Answer Post are highlighted in \textcolor{red}{red}, which are not in the query but exist only in the metadata and answer posts. Additionally, among those notable terms, some terms that are not covered by the query expansion approach are further highlighted in \textcolor{red}{\textbf{bold}}.}
\vspace{-0.1in}
\label{tab:case_study}
\scriptsize
\centering
\resizebox{\textwidth}{!}{
\renewcommand{\arraystretch}{1.1}
\begin{tabular}{l c}
\toprule
\multicolumn{2}{c}{\textbf{StackExchange-Any Answer w/ SplitByTime}} \\ \midrule
\textbf{Query} & \multicolumn{1}{p{.9\textwidth}}{\textsf{\textbf{[Title]}} Choosing activation and loss functions in autoencoder \newline
\textsf{\textbf{[Text]}} I am following this keras tutorial to create an autoencoder using the MNIST dataset. Here is the tutorial: \textsf{<URL>}. However, I am confused with the choice of activation and loss for the simple one-layer autoencoder (which is the first example in the link). Is there a specific reason sigmoid activation was used for the decoder part as opposed to something such as relu? I am trying to understand whether this is a choice I can play around with, or if it should indeed be sigmoid, and if so why? Similarily, I understand the loss is taken by comparing each of the original and predicted digits on a pixel-by-pixel level, but I am unsure why the loss is binary crossentropy as opposed to something like mean squared error. I would love clarification on this to help me move forward! Thank you!} \\
\noalign{\vskip 0.25ex}\cdashline{1-2}\noalign{\vskip 0.75ex}
\textbf{MetaData} & 
\multicolumn{1}{p{.9\textwidth}}
{\textsf{\textbf{[comments in answers by pid]}}:
\([\){"I wrote about it here, but it was ages ago so I cannot find it now; \textcolor{red}{BCE}'s properties as a function means it's not the best choice for image data, even in greyscale. Unlike \textcolor{red}{\textcolor{red}{MSE}}, it is asymmetrically biased against overconfidence, so it systematically underestimates the values, needlessly dimming the output intensities. And, as this question shows, causes unnecessary confusion on top."},\newline
{"Hmm. I think you may be correct in general, but for this particular use case (an autoencoder), it's been empirically and mathematically shown that training on the \textcolor{red}{BCE} and \textcolor{red}{MSE} objective both yield the same optimal reconstruction function: \textsf{<URL>} — but that's just a minor detail."},\newline
{"I cannot load the pdf for some reason, but I'm not surprised - the minima of both losses are the same if your goal is to autoencode a 1:1 match of intensities. It's just not always an optimal loss if your goal is to have a nice-looking image; e.g. MNIST would probably look best with most pixels being either 1 or 0 (in/not in the set of pixels for the character, basically learning a topology)."}\(]\),\newline
\textsf{\textbf{[tags by pid]}}: \([\)`neural-networks', {`loss-functions'}, `\textcolor{red}{\textbf{keras}}', {`autoencoders'}\(]\)
} \\ \noalign{\vskip 0.25ex}\cdashline{1-2}\noalign{\vskip 0.75ex}
\multicolumn{1}{p{.1\textwidth}}{\textbf{Answer Post}} & \multicolumn{1}{p{.9\textwidth}}{You are correct that \textcolor{red}{MSE} is often used as a loss in these situations. However, the \textcolor{red}{\textbf{Keras}} tutorial (and actually many guides that work with MNIST datasets) normalizes all image inputs to the range \([\)0, 1\(]\). This occurs on the following two lines: \texttt{x\_train = x\_train.astype(float32) / 255, x\_test = x\_test.astype(float32) / 255}. Note: as grayscale images, each pixel takes on an intensity between 0 and 255 inclusive. Therefore, \textcolor{red}{BCE} loss is an appropriate function to use in this case. Similarly, a sigmoid activation, which squishes the inputs to values between 0 and 1, is also appropriate. You'll notice that under these conditions, when the decoded image is "close" to the encoded image, \textcolor{red}{BCE} loss will be small. I found more information about this \textsf{<URL>}.} \\ \midrule
\textbf{Retrieval Rank} & \multicolumn{1}{p{.9\textwidth}}{\small{No Expan. : 26\qquad
Expan. w/ Full :  15\qquad
Expan. w/ Retriever :  38\qquad
DAQu (Ours) : 6}} \\ \midrule
\multicolumn{2}{c}{\textbf{StackExchange-Best Answer w/ SplitByTime}} \\ \midrule
\textbf{Query} & 
\multicolumn{1}{p{.9\textwidth}}{\textsf{\textbf{[Title]}} When to Normalization and Standardization?
\newline \textsf{\textbf{[Text]}} 
I see pro-processing with Normalization, which aligns data between 0 and 1, and standardization makes zero mean and unit variance. And multiple standardization techniques follow on.. Any clear definition at what cases what should be used? Thanks in Advance!!
} \\
\noalign{\vskip 0.25ex}\cdashline{1-2}\noalign{\vskip 0.75ex}
\textbf{MetaData} & 
\multicolumn{1}{p{.9\textwidth}}{\textsf{\textbf{[comments]}}: [`hi @onestop, is it ok to take \textcolor{red}{\textbf{log transformation}} only to skewed columns?']\newline
\textsf{\textbf{[current tags]}}:[`normalization', `feature-scaling']
} \\ \noalign{\vskip 0.25ex}\cdashline{1-2}\noalign{\vskip 0.75ex}
\textbf{Answer Post} & \multicolumn{1}{p{.9\textwidth}}{
In unsupervised learning, the scaling of the features has a great influence on the result. If a feature has a variance that is many times greater, it can dominate the target function of the algorithm. Therefore, it is of great importance to scale the input data in a way that their variability matches or at least does not contradict the semantics. There are several \textcolor{red}{\textbf{transformation methods}} to put the features into a comparable form. These use different forms of normalization or standardization according to their context. (...) 
} \\ \midrule
\textbf{Retrieval Rank} & \multicolumn{1}{p{.9\textwidth}}{\small{No Expan. : 244\qquad
Expan. w/ Full :  178\qquad
Expan. w/ Retriever :  347\qquad
DAQu (Ours) : 105}} \\ \midrule
\multicolumn{2}{c}{\textbf{Amazon-Future Purchase w/ ProductToProduct }} \\ \midrule
\textbf{Query} & 
\multicolumn{1}{p{.9\textwidth}}{Kindergarten-Grade 3. Fox has composed a simple refrain to celebrate human connections in this lovely picture book. ``Little one, whoever you are,'' she explains, there are children all over the world who may look different, live in different homes and different climates, go to different schools, and speak in different tongues but all children love, smile, laugh, and cry. Their joys, pain, and blood are the same, ``whoever they are, wherever they are, all over the world.'' Staub's oil paintings complement the simple text. She uses bright matte colors for the landscapes and portraits, placing them in gold borders, set with jewels and molded from plaster and wood. These frames enclose the single- and double-page images and echo the rhythm of the written phrases. Within the covers of the book, the artist has created an art gallery that represents in color, shape, and texture, the full range of human experience.
} \\
\noalign{\vskip 0.25ex}\cdashline{1-2}\noalign{\vskip 0.75ex}
\textbf{MetaData} & 
\multicolumn{1}{p{.9\textwidth}}{
\textsf{\textbf{[previous product description]}}:[
``Betsy Snyder's first board book as an author-illustrator, <em>Haiku Baby</em> follows a tiny bluebird, the book's would-be protagonist, as it visits its various animal companions--from an elephant that shades the bird with a parasol to a fox in a meadow and a whale in the ocean. The little bird's story is told primarily in pictures, and through the book's six haiku: rain, flower, sun, leaf, snow, and--of course, it would not be a board book without--the moon, making it ideal for the bedtime line-up. Adorable collage-cut illustrations work nicely with the haiku form to give the book a whimsical, yet serene, feel. And the haiku are light and fun without being too cutesy. Index tabs on the right margin, with pictures that tie to each of the poems (leaf, raindrop, snowflake, etc.), create a unique look, and make it easy for toddlers to flip through the pages on their own without having them stick together like they can with other board books. Snyder excels at visual storytelling and short forms, possibly a talent she honed as a designer/illustrator in the kids' greeting card business. In the world of board books, this slender little volume really stands out''
]\newline
\textsf{\textbf{[previous product category]}}:[
``Books'', ``\textcolor{red}{\textbf{Children's Books}}'', ``Early Learning''
]\newline
\textsf{\textbf{[previous review text]}}:[
``\textcolor{red}{My baby loves this book}. It has been mouthed, pulled, and thrown many times and still looks new. No tears or running on the pages. No words inside, but has the song on the back incase one does not know it. Can easily make your own story up. My sister washed her book, which you should not do, and it got wrinkled and looks worn down. It did not tear or come apart though'', \newline
`Nice little book. Has all the seasons and some weather.'
]
} \\ \midrule
\textbf{Future Product} & \multicolumn{1}{p{.9\textwidth}}{
\textsf{\textbf{[Title]}} Ten Little Fingers and Ten Little Toes
\newline \textsf{\textbf{[Text]}} 
``There was one little baby who was born far away. And another who was born on the very next day. And both of these babies, as everyone knows, had ten little fingers and ten little toes." So opens this nearly perfect picture book. Fox's simple text lists a variety of pairs of babies, all with the refrain listing the requisite number of digits, and finally ending with the narrator's baby, who is 11truly divine'' and has fingers, toes, 11and three little kisses/on the tip of its nose.'' Oxenbury's signature multicultural babies people the pages, gathering together and increasing by twos as each pair is introduced. They are distinctive in dress and personality and appear on primarily white backgrounds. The single misstep appears in the picture of the baby who was ``born on the ice.'' \textcolor{red}{\textbf{The child}}, who looks to be from Northern Asia or perhaps an Inuit, stands next to a penguin. However, this minor jarring placement does not detract enough from the otherwise ideal marriage of text and artwork to prevent the book from being a first purchase. Whether shared one-on-one or in storytimes, where the large trim size and big, clear images will carry perfectly, this selection is sure to be a hit.''
} \\ \midrule
\textbf{Retrieval Rank} & \multicolumn{1}{p{.9\textwidth}}{\small{No Expan. : 29\qquad
Expan. w/ Full :  162\qquad
Expan. w/ Retriever :  765\qquad
DAQu (Ours) : 27}} \\ \bottomrule
\bottomrule

\end{tabular}
}
\end{table*}
\begin{table*}[t!]
\caption{Error case study for StackExchange (Best Answer).}
\vspace{-0.1in}
\label{tab:error_case_study}
\scriptsize
\centering
\resizebox{\textwidth}{!}{
\renewcommand{\arraystretch}{1.1}
\begin{tabular}{l c}
\toprule
\multicolumn{2}{c}{\textbf{StackExchange-Best Answer w/ SplitByUser}} \\
\midrule
\textbf{Query} & 
\multicolumn{1}{p{.9\textwidth}}{
\textsf{\textbf{[Title]}} How can I determine accuracy of past probability calculations? \newline
\textsf{\textbf{[Text]}} I do not study statistics but engineering, but this is a statistics question, and I hope you can lead me to what I need to learn to solve this problem. I have this situation where I calculate probabilities of 1000's of things happening in like 30 days. If in 30 days I see what actually happened, how can I test to see how accurately I predicted? These calculations result in probabilities and in actual values (ft). What is the method for doing this? Thanks, CP
} \\
\noalign{\vskip 0.25ex}\cdashline{1-2}\noalign{\vskip 0.75ex}
\textbf{MetaData} & 
\multicolumn{1}{p{.9\textwidth}}{
\textsf{\textbf{[answer posts]}}: [`You should check out "area51.stackexchange.com/proposals/117/quantitative-finance?referrer=b3Z9BBygZU6P1xPZSakPmQ2"> area51.stackexchange.com/proposals/117/quantitative-finance?referrer=b3Z9BBygZU6P1xPZSakPmQ2", they are trying to start one on stackexchange.com.'] \newline
\textsf{\textbf{[comments]}}: [`The Federalist paper is very interesting and describes similar to above answer. Thanks.'] \newline
\textsf{\textbf{[current tags]}}: [`probability']
} \\
\noalign{\vskip 0.25ex}\cdashline{1-2}\noalign{\vskip 0.75ex}
\textbf{Answer Post} & 
\multicolumn{1}{p{.9\textwidth}}{
What you're looking for are called Scoring Rules, which are ways of evaluating probabilistic forecasts. They were invented in the 1950s by weather forecasters, and there has been some work on them in the statistics community. One thing you could do would be to bin the forecasts by probability range (e.g.: 0--5\%, 5\%--10\%, etc.) and look at how many predicted events in that range occurred. If there are 40 events in the 0--5\% range and 20 occur, then you might have problems. If the events are independent, then you could compare these numbers to a binomial distribution. (...)
} \\
\midrule
\textbf{Retrieval Rank} & 
\multicolumn{1}{p{.9\textwidth}}{
\small{Expan. w/ Full: 24 \qquad DAQu (Ours): 4}
} \\
\bottomrule
\end{tabular}
}
\end{table*}

\end{document}